\documentclass[10pt,letterpaper,twocolumn]{article}
\usepackage[margin=0.75in,top=1in,bottom=1in,columnsep=0.375in]{geometry}
\usepackage{newtxtext}
\usepackage{helvet}
\usepackage{courier}
\usepackage[hyphens]{url}
\usepackage{graphicx}
\usepackage[round,authoryear]{natbib}
\usepackage{caption}
\usepackage[caption=false,font=small]{subfig}
\usepackage{booktabs}
\usepackage{amsmath,amssymb}
\usepackage{tabularx}
\usepackage{placeins}
\usepackage{flafter}
\usepackage{needspace}
\usepackage{titlesec}
\titleformat{\section}{\large\bfseries\centering}{}{0pt}{}
\titleformat{\subsection}{\normalsize\bfseries}{\thesubsection}{0.6em}{}
\titleformat{\subsubsection}[runin]{\normalsize\bfseries}{\thesubsubsection}{0.6em}{}[.]
\newcommand{\RACE}{RACE}

\newcommand{\metric}[1]{\mbox{#1}}
\renewenvironment{abstract}{%
  \centerline{\normalsize\bfseries\abstractname}%
  \vspace{0.5ex}%
  \setlength{\leftmargini}{10pt}%
  \begin{quote}%
    \fontsize{9}{10}\selectfont%
}{%
  \par\end{quote}%
  \vskip 1ex%
}
\title{\Large\bfseries RACE: Residual-Aware Test-Time Adaptation for Neighbor-Rich Time-Series Foundation Model Forecasting}
\author{\normalsize Hao-Nan Shi\thanks{Equal contribution.}\quad Tong Wu\footnotemark[1]\quad Chen-Cong Sun\\
\normalsize Yuan Jiang\quad Han-Jia Ye\quad De-Chuan Zhan\thanks{Corresponding author: \texttt{zhandc@nju.edu.cn}.}\\[0.6em]
\small School of Artificial Intelligence, Nanjing University\\
\small National Key Laboratory for Novel Software Technology, Nanjing University\\
\small Nanjing, China}
\date{}
\begin{document}
\maketitle

\begin{abstract}
Time-series foundation models (TSFMs) perform strongly across forecasting tasks, but their per-series inference is ill-suited to \emph{neighbor-rich forecasting}, where each query has access to related but nonidentical historical series. Continuous glucose monitoring (CGM) and Web/cloud workloads exemplify this setting: CGM trajectories share physiological patterns but vary across individuals, devices, and conditions, while Web/cloud workloads combine common operating regimes with non-stationarity, heavy tails, and bursts. Such histories contain useful shared structure, yet individual neighbors are not equally relevant. Existing methods either fine-tune TSFMs for each target domain, incurring additional costs and offering limited transferability across backbones, or append retrieved series without verifying whether they support the current forecast. The key challenges are conflicting residual evidence from neighboring series and residual patterns that vary across TSFMs and forecasting tasks.
We formulate test-time neighborhood scaling: we directly leverage same-domain neighbor evidence without modifying the backbone. We propose RACE (Residual-Aware Correction of Forecasting Errors), a two-stage framework for selecting and utilizing historical neighbors. We first retrieve query-compatible neighbors, align their residuals to the query scale, and aggregate coherent evidence into the training-free RACE-TF correction. Full RACE then uses a lightweight, domain-specific Gate to determine when applying the correction is beneficial, with a training workflow that can be reused across TSFM backbones. Across four TSFMs, RACE improves all three domain-aggregate metrics on both primary domains, with the largest gains on high-error queries. Within each domain, a Gate trained on one TSFM transfers to other backbones without adaptation, and the resulting pipeline improves all 72 cross-backbone metric comparisons over the matched frozen targets.
\end{abstract}

\begin{figure*}[t]
  \centering
  \includegraphics[width=\textwidth,keepaspectratio]{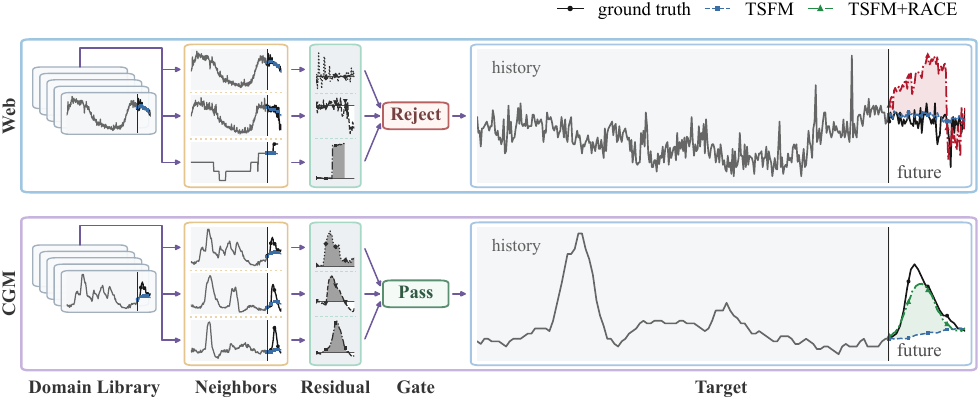}
  \caption{Neighbor-rich forecasting turns reusable domain histories into selective corrections. Each row uses outcome-observed windows and frozen-TSFM residuals from the same domain. Conflicting Web evidence is rejected, whereas coherent CGM neighbors improve the forecast.}
  \label{fig:evidence}
\end{figure*}

\section{Introduction}

The growing capability of time-series foundation models (TSFMs) is shifting forecasting from task-specific training toward broad model reuse. Chronos, TimesFM, and Moirai established zero-shot forecasting through different generative and universal-training designs \citep{ansari2024chronos,das2024timesfm,woo2024moirai}, while Chronos-2 and Toto-2.0 further extend model coverage and accuracy \citep{ansari2025chronos2,khwaja2026toto}. In the per-series deployment setting studied here, each query is forecast from its own context. Related historical series already available in the deployment domain remain unused by this inference route.

We study this overlooked structure through \emph{neighbor-rich forecasting}. As illustrated in Figure~\ref{fig:evidence}, each query is associated with a reusable domain library of related but distinct historical series or windows. Continuous glucose monitoring (CGM) and Web/cloud workload forecasting provide two representative examples. CGM systems accumulate trajectories that share physiological patterns but vary across patients, sensors, and conditions. Web/cloud systems similarly retain histories from related workloads and services that share operating regimes while exhibiting non-stationarity, heavy tails, and bursts. Related histories already inform health applications: case-based diabetes systems retrieve prior patient cases \citep{montani2000diabetic}, and wearable stress models retrieve historical patterns for personalization \citep{simon2026wearable}. These examples motivate using neighbors as evidence, while leaving open which residual corrections a frozen forecaster should trust. Once the outcome of a historical window is observed, the difference between the TSFM forecast and the realized outcome forms a model-specific residual record. Such records capture how the model has erred in related contexts and can provide local evidence for the current forecast. As the contrasting cases in Figure~\ref{fig:evidence} show, compatible neighbors may support a useful correction, whereas conflicting ones should be rejected.

Retrieval-augmented methods reuse domain libraries at different stages: learned retrieval or fusion modules integrate neighboring sequences \citep{ning2025tsrag,han2025raft,zhang2024timeraf}, while training-free input augmentation supplies historical examples to frozen forecasters \citep{tire2024raf,yang2025timerag}. Prediction fusion also supports neighbor-based personalization without retraining \citep{indriani2026neighbor}. These approaches motivate a more specific question: do similar histories produce residuals that support the same correction? Neighbor residuals may conflict, and their patterns vary across TSFMs and forecasting tasks. Direct averaging can therefore degrade the forecast, while a backbone-specific fusion interface can complicate reuse across models.

We formulate this problem as \emph{test-time neighborhood scaling}: using a domain library as a neighborhood of temporally valid, same-domain evidence to improve a query forecast without updating the TSFM backbone. We propose \RACE{} (Residual-Aware Correction of Forecasting Errors), a two-stage framework that separates evidence construction from correction selection. Residual Evidence Construction retrieves query-compatible neighbors using both the query context and the frozen forecast, aligns their residuals to the query scale, and aggregates consistent evidence. The resulting consensus-mean proposal defines the \RACE-TF{} variant, where TF denotes training-free. Full \RACE{} adds a lightweight domain-specific Gate that uses inference-visible information to decide whether to apply the correction. The Gate is fitted on development data and frozen at deployment; test-time adaptation here changes the forecast output, not the backbone parameters. The same fitting procedure is used across TSFMs, and the transfer experiments additionally reuse a fitted source Gate and its threshold.

We evaluate \RACE{} using four TSFMs in two application domains, CGM and Web/cloud, each drawing on multiple datasets. For reproducibility, CGM performance is evaluated on the public dataset, with the private collections used only as retrieval resources. Our analyses cover forecasting accuracy, mechanism behavior, efficiency, and cross-backbone transfer. \RACE{} improves all three domain-aggregate metrics in both primary domains across the four backbones, with the largest gains on queries for which the frozen backbone has high prediction error. Further analyses show that residual evidence construction remains effective without target-domain training and selective application improves the reliability of candidate corrections. A source-trained Gate can also be reused on other TSFMs within the same domain without target-specific fitting or recalibration: all 72 cross-backbone metric comparisons improve on the matched frozen targets, with average gains close to those of target-trained RACE.

Our contributions are threefold. We define \emph{neighbor-rich forecasting} and formulate the use of a reusable domain library as \emph{test-time neighborhood scaling}. We introduce \RACE{}, which constructs a consensus residual correction and selectively applies it; its evidence-construction stage also yields the training-free \RACE-TF{} variant. Finally, experiments in the CGM and Web/cloud domains across four TSFMs show domain-aggregate forecasting gains, particularly on high-error queries, and demonstrate that trained Gates can be reused across backbones without target-specific fitting or recalibration.

\section{Related Work}

\noindent\textbf{Time-series foundation models and deployment residuals.}
Time-series foundation models extend zero-shot forecasting through broad pretraining, with probabilistic language-style modeling in Chronos \citep{ansari2024chronos}, decoder-only modeling in TimesFM \citep{das2024timesfm}, universal training in Moirai \citep{woo2024moirai}, and more recent universal or scaling-oriented families in Chronos-2 and Toto-2.0 \citep{ansari2025chronos2,khwaja2026toto}. GIFT-Eval, a benchmark for general time-series forecasting, makes cross-domain, leakage-aware evaluation explicit \citep{aksu2024gift}. RACE is not another backbone. It studies a post-deployment question left open by a frozen backbone: whether its own legally observed residual history can refine a later output under a disciplined information boundary. 

\noindent\textbf{Retrieval and post-hoc forecasting correction.}
Retrieval-based forecasting uses related historical sequences as references \citep{jing2022retime}. Retrieval Augmented Forecasting (RAF) retrieves examples for training-free input augmentation \citep{tire2024raf}. TimeRAG uses dynamic time warping (DTW) to retrieve sequences for a large language model (LLM) prompt \citep{yang2025timerag}. TimeRAF is a retrieval-augmented foundation model with learned retrieval and integration \citep{zhang2024timeraf}; TS-RAG applies retrieval-augmented generation to time series through an adaptive fusion module with a frozen backbone \citep{ning2025tsrag}. Retrieval Augmented Time Series Forecasting (RAFT) instead learns to combine retrieved contexts and futures during forecasting-model training \citep{han2025raft}. Context-based calibration adapts a prediction head using context-similar samples \citep{chen2024calibration}. These methods operate on sequence content or learned predictors; \RACE{} aligns and filters the frozen forecaster's own historical residuals before deciding whether to modify its output.

Optimization-Free Test-Time Adaptation (OFTTA) adapts normalization and support-set prototypes for wearable activity recognition without backpropagation \citep{wang2023oftta}. Such adaptation is distinct from \RACE{}'s output-side residual correction. In full \RACE{}, the Gate is fitted before deployment; only the evidence-construction variant is training-free.

\noindent\textbf{Case-based prediction and personalization in health.}
Case-based glucose prediction estimates future glucose from similar historical trajectories and studies how distance measures, neighbor counts, and aggregation affect accuracy \citep{zulj2021data}. Neighbor Fusion Personalization combines a pretrained population predictor with population and personal neighbor estimates, adjusting their contribution as personal history accumulates without retraining \citep{indriani2026neighbor}. GlyRAG combines glucose-derived textual context and retrieved episodes through learned retrieval adapters and fusion; its LLM-generated descriptions can be cached offline \citep{soumma2026glyrag}. For wearable stress detection, frozen foundation-model embeddings support retrieval from a user's history and personalization of downstream representations \citep{simon2026wearable}. These studies establish the value of historical cases and lightweight personalization. \RACE{} addresses a specific correction object: residuals produced by the same frozen forecaster on completed windows. It aligns and filters these residuals into a consensus proposal, then uses a utility-trained Gate to apply the proposal or retain the base output. Our experiments evaluate this residual-correction mechanism rather than claim a performance comparison with these health-specific systems.

\section{Problem Definition and Challenges}

\subsection{From Time-Series to Neighbor-Rich Forecasting}

A standard forecasting task maps a query window's observed context to its unobserved future. Under the per-series inference protocol considered here, a frozen TSFM does so without consulting completed records in the deployment environment. Here, we consider \emph{neighbor-rich forecasting}, in which a target query is accompanied by a library of related but non-identical series. Completed windows provide legitimately available realized futures, frozen-model forecasts, and residuals; these outcome-observed historical windows are reusable evidence. Continuous glucose monitoring (CGM) and Web/cloud workload forecasting are representative settings. CGM trajectories share physiological dynamics but differ by individual, device, and state; workload series share operating regimes but exhibit non-stationarity, bursts, and heavy tails. Their libraries contain useful common structure, but no individual neighbor is directly interchangeable with the target. While this is our primary setting, Appendix~E.1 reports 34 GIFT-Eval subsets meeting the fixed-window requirements without a predefined narrow neighbor relation, where the same legal retrieval and residual-evidence process can still identify reusable windows.

\subsection{In-Domain Residual Adaptation}

A frozen TSFM learns broad temporal regularities from large-scale, multi-domain pretraining, yet per-series inference without retrieval ignores both completed related windows in the deployment domain and its realized errors on them. This domain library is already available and requires no backbone update. This enables output-side correction: when a completed window and a query are comparable in temporal and frozen-model forecasting behavior, the former's residual can inform the latter's correction. The TSFM remains frozen, and only earlier outcome-observed domain records modify its output.

\subsection{Residual-Correction and Information Boundary}

For a query window $q$, let $\mathbf c_q$ be its observed context and
$\mathbf y_q$ its future vector. A frozen forecaster $f_\theta$ produces the base forecast $\hat{\mathbf y}_q$. A historical window $i$ becomes a residual record only after its future has been observed:
\begin{equation}
\begin{gathered}
\hat{\mathbf y}_q=f_\theta(\mathbf c_q),\qquad
\hat{\mathbf y}_i=f_\theta(\mathbf c_i),\\
\mathbf r_i=\mathbf y_i-\hat{\mathbf y}_i.
\end{gathered}
\label{eq:base-residual}
\end{equation}
Here, $i$ denotes a completed historical window whose future is already
observed; $\mathbf y_i$ and $\hat{\mathbf y}_i$ are its observed future and
frozen-model forecast, respectively. In contrast, $\mathbf y_q$ is
unavailable when the correction for $q$ is constructed. Residual correction
uses the legal library available before $q$ to construct a dimensionless
proposal $\boldsymbol\Delta_q$ and decide whether to apply it:
\begin{equation}
\hat{\mathbf y}_q^{\mathrm{final}}
=\hat{\mathbf y}_q+a_qs_q\boldsymbol\Delta_q,
\qquad a_q\in\{0,1\}.
\label{eq:residual-action}
\end{equation}
Here, $s_q$ is the mean absolute value of the observed entries in the query
context, floored at $10^{-6}$. Each neighbor residual is first normalized by
its own history scale and aligned to the query by the clipped ratio
$\operatorname{clip}(s_i/s_q,1/4,4)$; hence
$\boldsymbol\Delta_q$ is dimensionless and $s_q\boldsymbol\Delta_q$ is in
the original units of the query forecast. Clipping guards against extreme scale ratios in either direction. For evaluation by the continuous ranked probability score (CRPS) \citep{gneiting2007proper}, the same additive term
$a_qs_q\boldsymbol\Delta_q$ is applied to every predicted quantile
trajectory, preserving the frozen TSFM's quantile ordering and spread.

\subsection{Challenges and Motivation}

A large domain library provides many candidate windows, but only some carry errors that are relevant to the current query. Similar observed trajectories can lead to different futures, and the same trajectory may expose different weaknesses in different frozen TSFMs. Retrieval must therefore account for both the visible context and the model's forecast, while excluding outcomes that were unavailable at prediction time.

Even among compatible neighbors, residuals can differ in scale or point in conflicting directions. A useful correction must align these records and identify a shared pattern rather than average every retrieved error. Agreement alone, however, does not guarantee that the correction will help an individual query. Together, these observations define three interrelated challenges: (i) effective neighbors are scarce within a large reusable library; (ii) retrieved information must be converted into a robust correction proposal; and (iii) even such a proposal can harm the forecast, so its reliability must be judged.

\section{RACE Method}

\begin{figure*}[t]
  \centering
  \includegraphics[width=\textwidth,keepaspectratio,trim=0 154bp 0 0,clip]{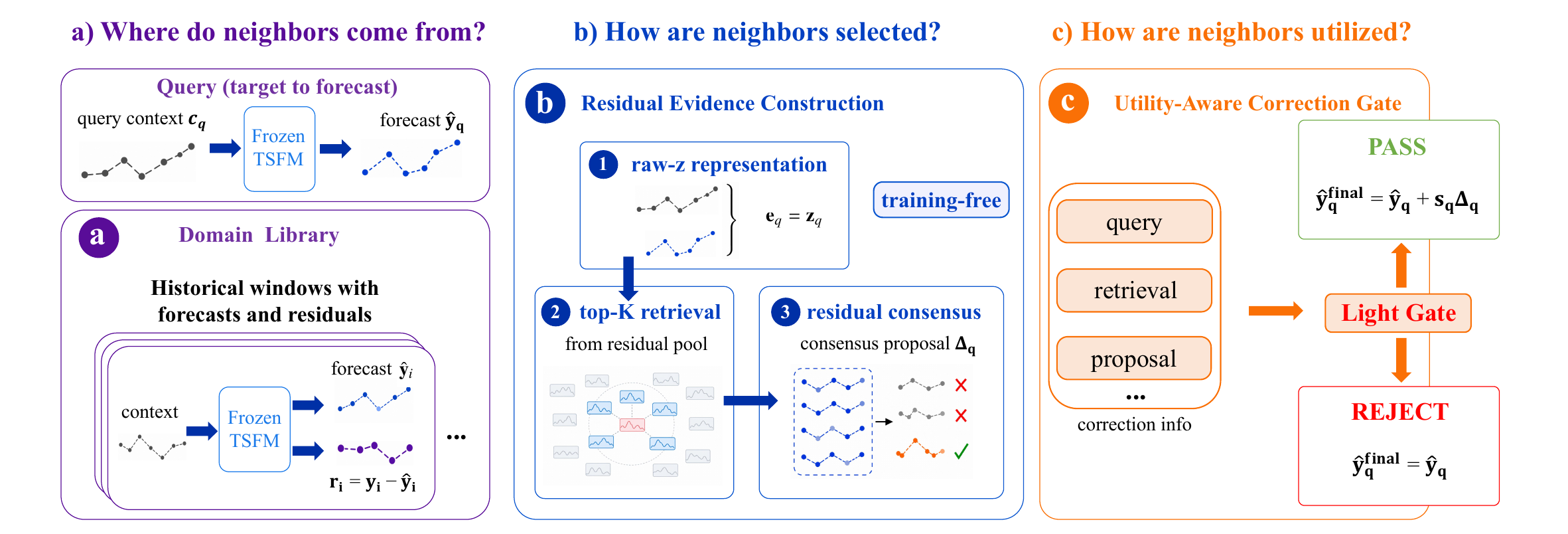}
  \caption{The pipeline of \RACE{}. A domain library supplies historical residual records; Residual Evidence Construction retrieves, aligns, and aggregates query-compatible neighbors, and the Utility-Aware Correction Gate decides whether to apply the proposal.}
  \label{fig:architecture}
\end{figure*}

As shown in Figure~\ref{fig:architecture}, \RACE{} draws on a reusable domain library and has two stages for selecting and using historical neighbor evidence. Residual Evidence Construction selects query-compatible neighbors, removes conflicting residuals, and aggregates the remainder into a training-free consensus correction; this stage alone is \RACE-TF{}. Since a proposal can still be beneficial, neutral, or harmful, the Utility-Aware Correction Gate learns when to use it.

\subsection{Residual Evidence Construction}

Residual evidence is informative only when a completed window resembles the query in both its observed trajectory and frozen-model forecast, because the residual reflects both. We therefore retrieve neighbors using a joint context–forecast descriptor rather than history alone.

Our default raw-z representation concatenates the query context and frozen forecast, both z-normalized using query-history statistics. Memory records retain the corresponding descriptors, residuals, scales, and provenance. \RACE{} retrieves $K=60$ eligible neighbors by squared Euclidean distance without accessing the query future, then aligns their residuals to the query scale using a clipped ratio. Residual consensus filters conflicting evidence before aggregation.

Our mean-centred residual-consensus filter uses median absolute deviation (MAD) for normalization and removes neighbors
whose query-scale-aligned residuals are displaced from, or more variable than,
the retrieved set. It forms a coordinate-wise mean centre over usable
coordinates. For neighbor $i$, $m_{qi}$ is its mean signed departure from that
centre and $v_{qi}$ is its within-neighbor population standard deviation after
removing this departure. The quantities $z^{(m)}_{qi}$ and $z^{(v)}_{qi}$ are,
respectively, the two-sided MAD-standardized displacement and the one-sided
MAD-standardized excess variation. With fixed robust threshold $\gamma$, the
retained-neighbor set is
\begin{equation}
\mathcal I_q^{(0)}
=
\left\{
i\in\mathcal V_q:
\left|z^{(m)}_{qi}\right|\le\gamma,\;
z^{(v)}_{qi}\le\gamma
\right\}.
\label{eq:consensus-filter}
\end{equation}
Here, $\mathcal V_q$ is the valid legal-neighbor set. If a MAD scale vanishes,
the implementation uses a standard-deviation fallback; if fewer than
$\lceil0.5|\mathcal V_q|\rceil$ neighbors pass screening, it retains the least
anomalous valid neighbors to that minimum. The retained query-scale-aligned
residuals are then averaged into the parameter-free consensus-mean proposal,
and the same mask is used for both the proposal and the Gate features. This
selection-and-aggregation pipeline is \RACE-TF{}; full \RACE{} selectively
acts on its unchanged proposal. Appendix~A gives the coordinate-level
definitions and fallback cases; Figure~\ref{fig:evidence-audit} audits the
resulting evidence quality.

\subsection{Utility-Aware Correction Gate}

A coherent consensus-mean proposal may still harm an individual query. We therefore train the Utility-Aware Correction Gate to estimate whether applying it will reduce mean squared error (MSE). For each labeled development window, the Gate target is
\begin{equation}
b_q=\mathbf{1}\!\left[U_{\mathrm{MSE}}(q,\boldsymbol\Delta_q)>0\right].
\label{eq:gate-action}
\end{equation}
Here,
$U_{\mathrm{MSE}}(q,\boldsymbol\Delta_q)
=
\operatorname{MSE}_{\mathcal H_q}(\hat{\mathbf y}_q,\mathbf y_q)
-
\operatorname{MSE}_{\mathcal H_q}
(\hat{\mathbf y}_q+s_q\boldsymbol\Delta_q,\mathbf y_q)$
is the MSE reduction from the always-apply correction on a labeled development
window. Thus, $b_q$ is available only during development. An XGBoost (XGB) gradient-boosted tree classifier \citep{chen2016xgboost} maps the
inference-visible feature vector $\mathbf x_q$ to $p_q$, and deploys
$a_q=\mathbf{1}[p_q\ge\tau]$, with $\tau$ frozen on calibration data. The
fixed 30-dimensional vector summarizes retained-evidence availability,
aligned-residual statistics, retrieval similarity, query and base-forecast
shape, proposal magnitude, weight concentration, and residual--proposal
alignment (Appendix~G), using the corrector's retained-neighbor, horizon, and
coordinate masks produced by Residual Evidence Construction.

An XGBoost classifier is fitted on the fit windows; model-select chooses the tree count by area under the precision--recall curve (PR-AUC); the classifier is then refitted on the combined fit and model-select windows with hierarchy-balanced weights; and calibration freezes $\tau$ by maximizing accepted MSE utility over attainable score quantiles. The test loader is created only afterward. The Gate neither changes the proposal nor retrains the frozen TSFM; it only decides whether to apply it.

All backbones share the same Gate feature schema, label, and chronological workflow. In-domain experiments fit a lightweight Gate per backbone because residuals are model-specific. The shared output-side features require no backbone-specific architecture or hidden-state adapter and allow direct cross-backbone Gate transfer, evaluated in Section~\ref{sec:cross-backbone-transfer}, with the full protocol in Appendix~E.2.

\section{Experimental Setup}
\label{sec:setup}

We evaluate \RACE{} in two neighbor-rich domains under a common chronological protocol, using four frozen backbones and matched forecasting comparisons.

\begin{table*}[t]
\caption{Main results across four frozen TSFMs. Base rows report raw errors; other rows report percentage improvements, with positive values indicating lower error. Best results are bolded. TS-RAG uses its supported Chronos-Bolt Base interface only. Comparisons use each backbone's matched base and test windows (Appendix~F).}
\label{tab:main-results}

\centering
\normalsize
\setlength{\tabcolsep}{3pt}
\renewcommand{\arraystretch}{1.0}
\begin{tabular}{@{}llrrc@{}}
\toprule
TSFM + correction & Method status & \multicolumn{1}{c}{Web: MSE / MAE / CRPS} & \multicolumn{1}{c}{CGM: MSE / MAE / CRPS} & Best Count \\
\midrule
Chronos-2 & frozen base & \metric{8.676e+05 / 70.35 / 51.27} & \metric{2.925 / 1.149 / 0.9099} & -- \\
+ RAF & training-free & \metric{\textbf{+2.503} / \textbf{+2.423} / +0.216} & \metric{+6.826 / +4.869 / +4.585} & 2 \\
+ RACE-TF (ours) & training-free & \metric{+1.159 / +0.099 / -0.186} & \metric{+20.752 / +10.531 / +9.849} & 0 \\
+ RACE (ours) & fitted (XGB) & \metric{+0.982 / +1.091 / \textbf{+1.182}} & \metric{\textbf{+20.814} / \textbf{+10.677} / \textbf{+9.939}} & \textbf{4} \\
\midrule
Chronos-Bolt Base & frozen base & \metric{9.277e+05 / 74.87 / 59.49} & \metric{3.170 / 1.233 / 0.9708} & -- \\
+ RAF & training-free & \metric{\textbf{+5.527} / +1.230 / +1.244} & \metric{+6.923 / +4.904 / +4.394} & 1 \\
+ TS-RAG & pretrained & \metric{-0.375 / -3.272 / -2.536} & \metric{+6.086 / +5.072 / +4.872} & 0 \\
+ TS-RAG & fine-tuned & \metric{-20.741 / -6.636 / -6.252} & \metric{+20.818 / \textbf{+14.526} / \textbf{+14.655}} & 2 \\
+ RACE-TF (ours) & training-free & \metric{+1.782 / +0.954 / +0.846} & \metric{+22.637 / +12.769 / +10.575} & 0 \\
+ RACE (ours) & fitted (XGB) & \metric{+1.647 / \textbf{+1.838} / \textbf{+1.945}} & \metric{\textbf{+22.773} / +12.928 / +10.688} & \textbf{3} \\
\midrule
TimesFM-2.5 & frozen base & \metric{7.358e+05 / 68.88 / 52.70} & \metric{3.151 / 1.178 / 0.9384} & -- \\
+ RAF & training-free & \metric{-0.125 / +0.431 / -2.010} & \metric{+5.453 / +1.268 / +1.923} & 0 \\
+ RACE-TF (ours) & training-free & \metric{\textbf{+2.342} / +0.845 / +1.033} & \metric{\textbf{+19.735} / +9.752 / +9.147} & 2 \\
+ RACE (ours) & fitted (XGB) & \metric{+1.986 / \textbf{+1.748} / \textbf{+2.187}} & \metric{+19.686 / \textbf{+9.927} / \textbf{+9.233}} & \textbf{4} \\
\midrule
Toto-2.0-313m & frozen base & \metric{6.616e+05 / 56.73 / 41.66} & \metric{3.243 / 1.186 / 0.9446} & -- \\
+ RAF & training-free & \metric{-5.266 / -3.827 / -6.165} & \metric{+18.197 / +10.108 / +10.168} & 0 \\
+ RACE-TF (ours) & training-free & \metric{\textbf{+1.301} / -0.126 / -0.719} & \metric{\textbf{+22.239} / \textbf{+11.281} / \textbf{+10.362}} & \textbf{4} \\
+ RACE (ours) & fitted (XGB) & \metric{+1.025 / \textbf{+1.047} / \textbf{+0.912}} & \metric{+22.032 / +11.268 / +10.311} & 2 \\
\bottomrule
\end{tabular}

\end{table*}

\subsection{Data and Temporal Protocol}

Web comprises six GIFT-Eval subsets \citep{aksu2024gift} and four Web/cloud workload subsets from the Large-scale Open Time Series Archive (LOTSA) \citep{woo2024moirai}. CGM uses the public ShanghaiT2DM collection \citep{zhao2023chinese} for target evaluation and four de-identified institutional collections solely as retrieval memory. Their completed windows supply historical residual evidence, but no Gate-fitting, model-selection, calibration, or test labels. The public CGM target collection is never an evidence source.

Every experiment uses a 512-step context and a 64-step forecast horizon. Future-label boundaries divide target series into fit (0--70\%), model-select (70--80\%), calibration (80--90\%), and test (90--100\%) sets. For Web, the union of fit windows from the ten logical subsets forms the residual pool; cross-series candidates are permitted, subject to the same-parent temporal rule and self-exclusion. CGM retrieves only from its four external collections. All test futures and residuals are excluded from retrieval, feature construction, threshold selection, and model selection. Appendices~B and C document source manifests, window sampling, overlap checks, privacy and release boundaries, and the leakage audit.

\subsection{Backbones, Comparators, and Evaluation}
\label{sec:backbone-evaluation}

We evaluate Chronos-2 Base (Chr.2) \citep{ansari2025chronos2}, Chronos-Bolt Base (ChrB.B) \citep{ansari2024chronosbolt}, TimesFM 2.5 200M (TFM.25) \citep{timesfm2025modelcard}, and Toto-2.0-313m (Toto2.313M) \citep{toto2026modelcard}; M denotes million parameters. Each backbone has independent caches of forecasts, residuals, representations, candidates, proposals, and Gate features; residuals are never shared across models. Defaults are raw-z retrieval, global legal memory, a consensus-mean proposal, $K=60$, and an MSE-targeted Gate. Standard experiments fit and calibrate one Gate per backbone; cross-backbone transfer reuses a source Gate and its frozen threshold (Section~\ref{sec:cross-backbone-transfer}).

Table~\ref{tab:main-results} compares \RACE{} with RAF and TS-RAG. RAF retrieves a Top-1 historical continuation and re-runs the frozen backbone on the augmented input without fitting. TS-RAG is evaluated only on its supported Chronos-Bolt Base interface, using either the pretrained (PT) or target-domain fine-tuned (FT) Adaptive Retrieval Mixer (ARM). Appendix~F records comparator implementations and checkpoint details.

For query $q$, $\mathcal H_q$ contains forecast coordinates within the emitted prediction length and observed in the held-out future. MSE, mean absolute error (MAE), and CRPS are computed per window over $\mathcal H_q$ and macro-averaged. Improvements are relative to each method's matched frozen base; positive values mean lower error. Appendices~E.2--E.3 report encoder and Gate-target sensitivity.

\section{Results}

{\widowpenalty=10000
The experiments address five connected questions: (1) whether \RACE{} improves end-to-end forecasting across frozen backbones and both domains; (2) whether query-conditioned retrieval and residual consensus improve residual-evidence quality; (3) whether the Utility-Aware Correction Gate enriches beneficial corrections while retaining useful coverage; (4) whether the gains can be explained by global bias or a history-only query; and (5) whether a Gate trained on one TSFM can retain useful corrections on other frozen backbones without target-specific fitting or recalibration.\par}

\begin{table*}[t]
\caption{Cross-backbone Gate transfer: equal-weight mean MSE/MAE/CRPS improvement (\%) over matched frozen targets. Diagonal values are target-trained RACE references from Table~\ref{tab:main-results}. Backbone variants follow Section~\ref{sec:backbone-evaluation}. Gate uplift excludes diagonal entries.}
\label{tab:cross-backbone-transfer}

\centering
\normalsize
\setlength{\tabcolsep}{3pt}
\renewcommand{\arraystretch}{1.08}
\begin{tabular}{@{}l*{10}{>{\normalsize}r}@{}}
\toprule
\normalsize Source Gate & \multicolumn{5}{c}{\normalsize Web/cloud} & \multicolumn{5}{c}{\normalsize CGM (ShanghaiT2DM target)} \\
\cmidrule(lr){2-6}\cmidrule(lr){7-11}
 & \small\shortstack{Chronos-2\\Base} & \small\shortstack{Chronos-Bolt\\Base} & \small\shortstack{TimesFM\\2.5 200M} & \small\shortstack{Toto-2.0-\\313m} & \small\shortstack{Gate\\uplift} & \small\shortstack{Chronos-2\\Base} & \small\shortstack{Chronos-Bolt\\Base} & \small\shortstack{TimesFM\\2.5 200M} & \small\shortstack{Toto-2.0-\\313m} & \small\shortstack{Gate\\uplift} \\
\midrule
Chronos-2 Base & +1.09 & +1.94 & +1.98 & +1.04 & +0.74 & +13.81 & +15.46 & +12.99 & +14.68 & +0.10 \\
Chronos-Bolt Base & +0.88 & +1.81 & +1.21 & +0.87 & +0.35 & +13.74 & +15.46 & +12.89 & +14.68 & +0.03 \\
TimesFM 2.5 200M & +1.09 & +1.93 & +1.97 & +1.02 & +0.78 & +13.63 & +15.44 & +12.95 & +14.47 & -0.04 \\
Toto-2.0-313m & +1.06 & +1.92 & +1.98 & +0.99 & +0.67 & +13.72 & +15.44 & +12.95 & +14.54 & +0.06 \\
\bottomrule
\end{tabular}

\end{table*}

\subsection{End-to-End Forecasting Improvements}

Table~\ref{tab:main-results} shows that \RACE{} and \RACE-TF{} together achieve the best results in 19 of the 24 backbone--task--metric comparisons. Full \RACE{} leads in 13 comparisons, substantially more than any individual comparator or variant. At the domain-aggregate level reported in Table~\ref{tab:main-results}, it improves all three metrics across all four backbones and both tasks. These gains come from post-processing frozen forecasts, without updating backbone parameters, introducing backbone-specific hidden-state adapters, or requiring an additional backbone forecast. Fine-tuned TS-RAG achieves the best CGM MAE and CRPS on its natively supported Chronos-Bolt Base backbone. RAF leads Web MSE on Chronos-2 and Chronos-Bolt Base and Web MAE on Chronos-2, while \RACE{} and \RACE-TF{} outperform it on every CGM metric across the four backbones.

\RACE{} keeps the backbone frozen and applies a lightweight output-side Gate without an additional backbone forecast. The same Gate interface is reusable across TSFMs. Retrieval uses the Facebook AI Similarity Search (FAISS) library; Appendix~I reports the four-backbone runtime measurements and their measurement boundaries.

\begin{table*}[t]
\caption{Mechanism ablation. History-only retrieval excludes the forecast; residual mean (Mean) averages unfiltered residuals. Global bias replaces local retrieval with a horizon-wise correction. Metrics and four-backbone averaging are defined in Section~\ref{sec:ablation}.}
\label{tab:ablation}

\centering
\normalsize
\setlength{\tabcolsep}{3pt}
\renewcommand{\arraystretch}{1.06}
\begin{tabular}{@{}llrr@{}}
\toprule
Variant & Change & Web (\%) & CGM (\%) \\
\midrule
RACE & None & \metric{+1.410 / \textbf{+1.431} / \textbf{+1.557}} & \metric{+21.327 / \textbf{+11.200} / \textbf{+10.043}} \\
\midrule
\multicolumn{4}{@{}l}{\textit{Panel A: evidence-construction ablations (Gate retained)}} \\
Global bias + Gate & Without local retrieval & \metric{-0.040 / -0.945 / -1.332} & \metric{+3.802 / +1.320 / +1.369} \\
History-only retrieval & Without forecast & \metric{+1.214 / +1.209 / +1.336} & \metric{+16.484 / +8.943 / +8.021} \\
Residual mean + Gate & Without consensus & \metric{\textbf{+2.247} / +0.984 / +0.917} & \metric{+21.716 / +10.154 / +9.086} \\
\midrule
\multicolumn{4}{@{}l}{\textit{Panel B: Gate ablation and combined control}} \\
RACE-TF (training-free) & Without Gate & \metric{+1.646 / +0.443 / +0.243} & \metric{+21.341 / +11.083 / +9.983} \\
Residual mean & Without consensus or Gate & \metric{+1.785 / -5.785 / -8.028} & \metric{\textbf{+22.325} / +9.939 / +9.040} \\
\bottomrule
\end{tabular}

\end{table*}

\subsection{Cross-Backbone Gate Transfer}
\label{sec:cross-backbone-transfer}

Within each domain, we transfer each source-trained Gate, with its feature schema and threshold fixed, to the other three TSFMs. Targets retain their own forecasts, residual pools, and consensus proposals, without Gate refitting, model selection, or recalibration. Two domains, four sources, three distinct targets, and three metrics yield $2\times4\times3\times3=72$ off-diagonal comparisons. Table~\ref{tab:cross-backbone-transfer} summarizes the results. Gate uplift averages the increment over target RACE-TF across three metrics and the three other targets within each domain, in percentage points.

The Gate adds considerably more on Web/cloud, consistently improving MAE and CRPS across source--target pairs. This is consistent with filtering harmful corrections in a variable, burst-prone setting, despite an MSE trade-off relative to RACE-TF. On CGM, RACE-TF already provides strong gains across backbones and metrics, leaving smaller, occasionally negative Gate increments. These results suggest that selection matters most when neighborhood evidence is less reliable. Source-trained Gates nevertheless preserve useful corrections across backbones without target-specific fitting or recalibration; Appendix~E.2 provides the metric-wise results.

\begin{figure*}[t]
  \centering
  \includegraphics[width=\textwidth,keepaspectratio]{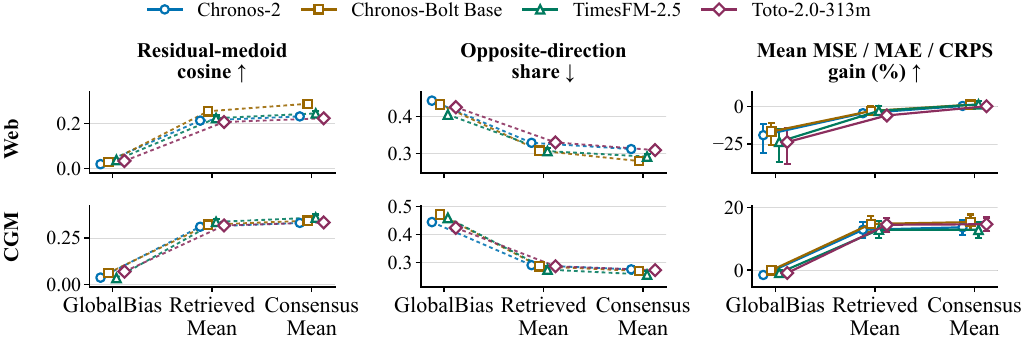}
  \caption{Residual-evidence construction audit. The first two columns show calibration-time mean residual-medoid cosine and the share of residuals with negative cosine to the medoid. The third shows post-hoc test gains averaged over MSE, MAE, and CRPS for ungated GlobalBias, Retrieved Mean, and Consensus Mean (RACE-TF). Error bars show 95\% target-parent bootstrap intervals.}

  \label{fig:evidence-audit}
\end{figure*}

\begin{figure*}[t]
  \centering
  \subfloat[Benefit enrichment]{\label{fig:action-audit}%
    \includegraphics[width=0.485\linewidth]{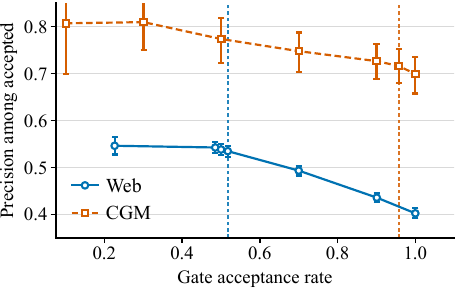}}
  \hfill
  \subfloat[Error-stratified outcomes]{\label{fig:error-stratified}%
    \includegraphics[width=0.485\linewidth]{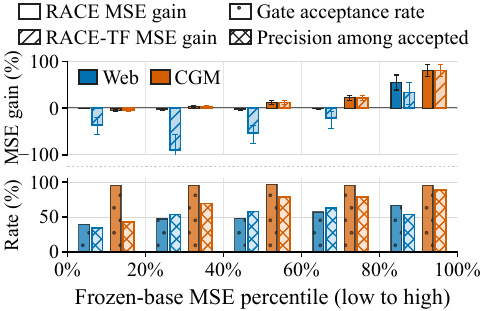}}
  \caption{Selective residual-transfer audit for Chronos-Bolt Base. (a) Test-set precision versus acceptance rate; dashed lines mark acceptance rates at calibration-fixed thresholds. (b) Five within-dataset base-MSE bins: correction gains (top), acceptance and precision (bottom). Error bars show 95\% target-parent bootstrap intervals.}
  \label{fig:selective-audit}
\end{figure*}

\begin{figure*}[t]
  \centering
  \includegraphics[width=\textwidth]{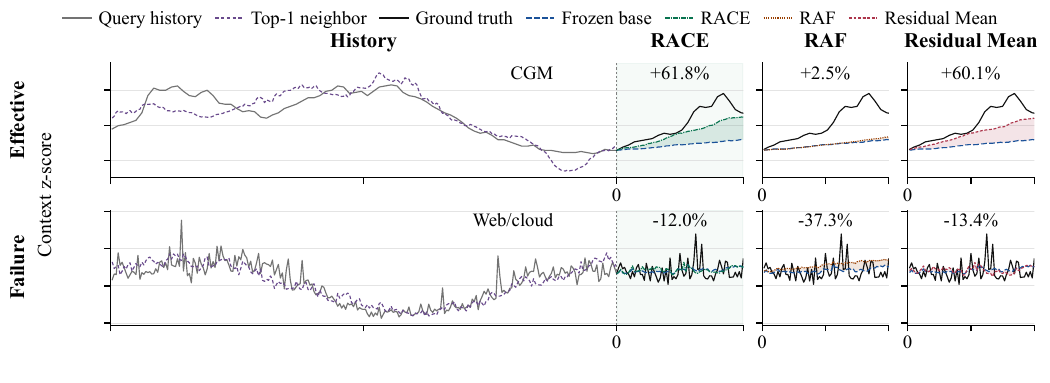}
  \caption{Chronos-2 case studies: CGM success (top) and Web failure (bottom). Left: the last 256 steps of the query and top-neighbor histories, followed by the 64-step RACE forecast window. Right: RAF and Residual Mean (without consensus or Gate) on the same future. Panels share the time-step scale and each row's vertical scale. Values are history-standardized; shading marks forecast changes. Percentages average MSE, MAE, and CRPS improvements.}
  \label{fig:case-study}
\end{figure*}

\subsection{Factorized Mechanism Ablation}
\label{sec:ablation}

Table~\ref{tab:ablation} reports MSE / MAE / CRPS percentage improvements over matched frozen bases, averaged equally across four backbones. Panel A retains the Gate while varying evidence construction; Panel B tests removing the Gate alone or together with consensus. These controls isolate query-conditioned retrieval, model-aware matching, residual consensus, and selective deployment.

In Panel A, \textbf{Global bias + Gate} replaces query-conditioned retrieval with a horizon-wise bias aggregated across queries. \textbf{History-only retrieval} removes the frozen forecast from the retrieval descriptor. \textbf{Residual mean + Gate} averages all retrieved residuals without residual consensus. The results show that global bias cannot represent local error structure, the forecast continuation helps identify residuals associated with the backbone's own failure mode, and consensus prevents conflicting neighbors from forming an unbalanced correction.

In Panel B, \RACE-TF{} applies the consensus proposal to every query, while Mean additionally removes residual consensus. These controls separate the effects of evidence filtering and selective deployment, showing why transferring an unfiltered neighborhood average is less reliable than combining query-conditioned retrieval, residual consensus, and gating. Appendix~E.3 also compares a direct learned correction control, the Consensus-Evidence Residual Regressor (CERR).

\subsection{Residual-Evidence Audit}

Figure~\ref{fig:evidence-audit} directly links evidence quality to correction utility. Replacing GlobalBias with query-conditioned retrieval increases residual directional agreement and reduces conflicting residuals across tasks and backbones. MAD-based consensus generally maintains or strengthens this agreement by removing inconsistent support. It improves the three-metric mean correction gain over Retrieved Mean on Web and for three of four CGM backbones; the TimesFM-2.5 CGM mean is nearly unchanged. The matching calibration-time coherence and final-test utility trends show that the gains arise from retrieving and integrating transferable error patterns rather than estimating a global shift.

\subsection{Selective Residual-Transfer Audit}

Figure~\ref{fig:selective-audit}\protect\subref{fig:action-audit} shows how the Gate enriches beneficial corrections through a precision--coverage trade-off: stricter thresholds retain fewer but more reliable corrections in both tasks. Calibration selects task-specific operating points, with stronger selectivity for Web and broader acceptance for CGM.

Figure~\ref{fig:selective-audit}\protect\subref{fig:error-stratified} shows that \RACE{} concentrates its largest gains on queries with high frozen-base error. Here, MSE gains use each dataset's mean base error as the denominator and are averaged equally across datasets (Appendix~E.3). The Gate reduces harmful corrections on easier Web queries while preserving substantial improvements in the high-error tails of both tasks. The high-error bins show strong gains, supporting the value of residual evidence for difficult forecasts. Appendix~E.3 reports the complete four-backbone results.

\Needspace{6\baselineskip}
\subsection{Qualitative Case Analysis}
\label{sec:case-study}

Figure~\ref{fig:case-study} examines individual Chronos-2 forecasts using \RACE{}, RAF, and Residual Mean, the ablation without consensus or the Gate. In the upper CGM case, the retrieved neighborhood supports a future glucose rise that the frozen forecast largely misses. Both \RACE{} and Residual Mean recover part of this rise, with \RACE{} giving a slightly larger improvement; RAF makes only a small adjustment and remains close to the underestimated baseline. In the lower Web case, the retrieved history follows the query's broad shape, but rapid fluctuations dominate the observed future. The smoother Chronos-2 forecast already tracks its local central level reasonably well, leaving little scope for a smooth residual adjustment; all three corrections instead worsen the forecast. This behavior highlights a difficulty also encountered when similar Web histories lead to different continuations: historical resemblance alone does not ensure transferable future or residual structure. Strong irregular fluctuations can further limit the benefit of either input augmentation or output correction. These observations suggest that the usefulness of a neighbor-rich library depends on the underlying data distribution. Reliable patterns shared across series, together with relatively weak irregular high-frequency variation, make the smoother CGM setting particularly favorable for neighbor-based enhancement.

\subsection{Robustness}

On the broader diagnostic over 34 eligible GIFT-Eval subsets, \RACE{} improves 11 of the 12 backbone--metric aggregates; Chronos-2 CRPS increases slightly, by 0.089\%. In cross-backbone transfer, the complete pipeline preserves most of the gains of target-trained correction and improves over matched frozen targets in $72/72$ domain--source--target--metric comparisons (Section~\ref{sec:cross-backbone-transfer}). Encoder, neighborhood-size, retrieval-scope, and Gate-target diagnostics show where the gains persist and where they weaken (Appendix~E). In particular, domain-level improvements do not imply improvements for every Web subset: Appendix~E.4 reports the corresponding win/loss counts.

\section{Conclusion}

RACE refines frozen-TSFM forecasts by aggregating compatible residuals from completed same-domain windows and selectively applying the correction through a lightweight utility gate. Across four TSFMs and two neighbor-rich domains, it delivers consistent domain-aggregate gains through lightweight post-processing, establishing historical model errors as reusable test-time evidence.

\FloatBarrier
\bibliographystyle{plainnat}
\bibliography{references}
\clearpage
\onecolumn
\appendix
\titleformat{\section}{\large\bfseries\centering}{\thesection}{0.6em}{}
\section*{Reader Navigation}

This technical supplement is organized as a self-contained implementation and evidence record. Appendix~A defines notation, scale alignment, and residual evidence construction. Appendix~B records data scope and release boundaries; Appendix~C fixes the chronological roles and eligible-residual-pool rules. Appendix~D describes the complete primary-result ledger. Appendix~E collects robustness, broader GIFT-Eval, and mechanism diagnostics. Appendices~F and~G specify comparator implementations and the Utility-Aware Correction Gate, respectively. Appendix~H records figure provenance and case selection. Appendix~I defines resource measurements and the availability boundary of the reproducibility materials. Appendix~J states the scope and capacity limitations of the present study; Appendix~K defines the backbone naming convention.

\section{Notation, Scales, and Residual Evidence Construction}

Table~\ref{tab:notation} is the authoritative notation inventory for the manuscript. It groups symbols that occur in the same computation, providing a compact map of legal retrieval, residual-evidence construction, and selective residual transfer. Bold symbols denote vectors, whose lengths depend on their roles; nonbold symbols denote scalars, indices, sets, operators, or actions. The table covers the core notation used in the manuscript.

\begin{table}[t]
\caption{Global mathematical notation used by \RACE{}.}
\label{tab:notation}
\begin{center}
\footnotesize
\setlength{\tabcolsep}{5pt}
\renewcommand{\arraystretch}{1.10}
\begin{tabular}{@{}p{0.36\textwidth}p{0.58\textwidth}@{}}
\toprule
Symbols & Meaning \\
\midrule
\textbf{Windows and forecasts}\newline $q,i;\ f_\theta$\newline $\mathbf c_q,\mathbf c_i;\ \mathbf y_q,\mathbf y_i$\newline $\hat{\mathbf y}_q,\hat{\mathbf y}_i,\hat{\mathbf y}_q^{\mathrm{final}}$ & $q$ is a query window and $i$ a historical window in $q$'s eligible residual pool. $f_\theta$ is the frozen time-series foundation model (TSFM); $\mathbf c_q,\mathbf c_i$ are visible contexts; $\mathbf y_q,\mathbf y_i$ are realized futures; and $\hat{\mathbf y}_q,\hat{\mathbf y}_i,\hat{\mathbf y}_q^{\mathrm{final}}$ are the base, neighbor, and deployed forecasts. \\
\textbf{Legal retrieval}\newline $\mathcal C_q,\mathcal N_q;\ K,P$\newline $t_i^{\mathrm{end}},t_q^{\mathrm{forecast\ start}}$ & $\mathcal C_q$ is the eligible residual pool and $\mathcal N_q$ its final subset of at most $K$ query-compatible neighbors. $P=600$ is the initial exact-ranking width; a short query is expanded exactly after legality filtering. A parent is an original full series; a same-parent candidate's future must end no later than the query future starts: $t_i^{\mathrm{end}}\le t_q^{\mathrm{forecast\ start}}$. \\
\textbf{Retrieval representations}\newline $\mathbf z_q;\ \phi_j,\operatorname{LN}_j,w_j$\newline $\operatorname{concat},\lVert\cdot\rVert_2,\langle\cdot,\cdot\rangle$\newline $\mathbf e_q,\mathbf e_i,\mathbf e_q^{\mathrm{HM}}$ & $\mathbf z_q$ is the normalized visible history--forecast trajectory. For $j\in\{1,\ldots,6\}$, $\phi_j$, $\operatorname{LN}_j$, and $w_j$ are feature extractor $j$, its layer-normalization transform, and its fixed weight. $\operatorname{concat}$ concatenates vectors, $\lVert\cdot\rVert_2$ is the Euclidean norm, $\langle\cdot,\cdot\rangle$ is the Euclidean inner product, and $\mathbf e_q,\mathbf e_i$ are query and neighbor representations; $\mathbf e_q^{\mathrm{HM}}$ is the HybridMath unit-norm representation. \\
\textbf{Residual scale alignment}\newline $\mathbf r_i;\ s_i,s_q$\newline $\widetilde{\mathbf r}_{qi};\ \operatorname{clip}(\cdot,\tfrac14,4)$ & $\mathbf r_i=\mathbf y_i-\hat{\mathbf y}_i$ is the realized neighbor residual. $s_i,s_q$ are mean absolute observed-context magnitudes, floored at $10^{-6}$; $\operatorname{clip}(x,\tfrac14,4)$ clamps $x$ to $[\tfrac14,4]$, and $\widetilde{\mathbf r}_{qi}$ is the dimensionless, clipped residual aligned to the query scale. \\
\textbf{Evidence sets}\newline $\mathcal V_q,\mathcal V_{qhc},\mathcal H_{qi}$\newline $\mathcal I_q^{(0)},\mathcal I_{qhc};\ h,c$ & $\mathcal V_q,\mathcal V_{qhc},\mathcal H_{qi}$ denote valid neighbors, coordinate-available neighbors, and usable coordinates. $\mathcal I_q^{(0)}$ and $\mathcal I_{qhc}$ are the initial and coordinate-specific retained sets; $(h,c)$ indexes a forecast-horizon coordinate and channel. \\
\textbf{Evidence summaries and screening}\newline $\bar r_{qhc},d_{qihc},m_{qi},v_{qi}$\newline $\mu_m,\mu_v,S(\cdot);\ c_{\mathrm{MAD}},\mathbf a,a_j$\newline $z^{(m)}_{qi},z^{(v)}_{qi},\gamma$ & $\bar r,d,m,v$ are the residual center, coordinate deviation, signed mean deviation, and within-neighbor standard deviation. $\mu_m,\mu_v,S$ are robust centers and median-absolute-deviation scales; $c_{\mathrm{MAD}}$ is the fixed MAD consistency coefficient; $\mathbf a$ is an across-neighbor scalar collection and $a_j$ one entry. $z^{(m)},z^{(v)}$ are MAD-standardized displacement and one-sided variability-excess scores; $\gamma$ is their fixed screening threshold. \\
\textbf{Proposal and selective transfer}\newline $\boldsymbol\Delta_q;\ \mathcal H_q;\ \operatorname{MSE}_{\mathcal H_q}$\newline $U_{\mathrm{MSE}}(q,\boldsymbol\Delta_q)$\newline $b_q,\mathbf x_q,p_q,\tau,a_q,\mathbf{1}[\cdot],w_i$ & $\boldsymbol\Delta_q$ is the query-normalized consensus-mean proposal (zero when no evidence remains). $\mathcal H_q$ is issued-and-observed forecast support, $\operatorname{MSE}_{\mathcal H_q}$ its per-window MSE operator, and $U_{\mathrm{MSE}}$ the RACE-TF improvement. $b_q$ is the development label; $\mathbf x_q$ is the inference-visible Gate feature vector; $p_q$, $\tau$, and $a_q=\mathbf{1}[p_q\ge\tau]$ are the probability, frozen threshold, and accepted-or-rejected action; $w_i$ is a valid-neighbor normalized proposal weight. \\
\textbf{Comparators and metrics}\newline $K_{\mathrm{RAF}};\ \ell_1,\ell_2$\newline $\mathrm{MSE},\mathrm{MAE},\mathrm{CRPS}$ & Retrieval Augmented Forecasting (RAF) uses $K_{\mathrm{RAF}}=1$ and prepends the selected historical context--future continuation to the query. The Consensus-Evidence Residual Regressor (CERR) is the direct learned residual-regression control in Appendix~E.3; it introduces no additional mathematical symbols. $\ell_1,\ell_2$ are XGBoost L1 and L2 regularizers. Mean squared error (MSE), mean absolute error (MAE), and continuous ranked probability score (CRPS) are computed per window on $\mathcal H_q$ and macro-averaged; a positive improvement means lower error than the matching frozen TSFM. \\
\bottomrule
\end{tabular}

\end{center}
\end{table}

The mean-centred, MAD-based residual consensus procedure is a \RACE{}-specific procedure, not a standard estimator with that name. It uses the conventional median absolute deviation (MAD) as a robust scale. Let $\mathcal V_q$ be the valid retrieved neighbors, $\mathcal H_{qi}$ the forecast coordinates available to neighbor $i$ after the evaluation and neighbor-horizon masks, and $\mathcal V_{qhc}\subseteq\mathcal V_q$ the neighbors available at coordinate $(h,c)$. The coordinate-wise centre and neighbor summaries are
\begin{equation}
\begin{gathered}
\bar r_{qhc}=\frac{1}{|\mathcal V_{qhc}|}
  \sum_{i\in\mathcal V_{qhc}}\widetilde r_{qihc},\\[0.35ex]
d_{qihc}=\widetilde r_{qihc}-\bar r_{qhc}.
\end{gathered}
\label{eq:supp-consensus-centre}
\end{equation}
\begin{equation}
\begin{gathered}
m_{qi}=\frac{1}{|\mathcal H_{qi}|}
  \sum_{(h,c)\in\mathcal H_{qi}}d_{qihc},\\[0.35ex]
v_{qi}=\left[\frac{1}{|\mathcal H_{qi}|}
  \sum_{(h,c)\in\mathcal H_{qi}}(d_{qihc}-m_{qi})^2\right]^{1/2}.
\end{gathered}
\label{eq:supp-consensus-summary}
\end{equation}
For an across-neighbor scalar collection $\mathbf a$, let $\operatorname{med}$ denote the median and $c_{\mathrm{MAD}}$ the fixed MAD consistency coefficient, and define
\begin{equation}
\begin{gathered}
\operatorname{MAD}(\mathbf a)=\operatorname{med}_j
  \left|a_j-\operatorname{med}_\ell a_\ell\right|,\\[0.35ex]
S(\mathbf a)=c_{\mathrm{MAD}}\,\operatorname{MAD}(\mathbf a).
\end{gathered}
\label{eq:supp-mad-scale}
\end{equation}
When $S(\mathbf a)$ is below the numerical tolerance, the implementation uses the population standard deviation of $\mathbf a$; when that also vanishes, the corresponding outlier score is zero. With $\mu_m=\operatorname{med}_{j\in\mathcal V_q}m_{qj}$ and $\mu_v=\operatorname{med}_{j\in\mathcal V_q}v_{qj}$, $S(m)$ and $S(v)$ below denote the robust scales of the respective values across $j\in\mathcal V_q$. The two scores and initial retained set are
\begin{equation}
\begin{gathered}
z^{(m)}_{qi}=\frac{|m_{qi}-\mu_m|}{S(m)},\\[0.35ex]
z^{(v)}_{qi}=\frac{v_{qi}-\mu_v}{S(v)},\\[0.35ex]
\mathcal I_q^{(0)}=\left\{i\in\mathcal V_q:\
  \left|z^{(m)}_{qi}\right|\le\gamma,\ z^{(v)}_{qi}\le\gamma\right\}.
\end{gathered}
\label{eq:supp-consensus-filter}
\end{equation}
The first score is two-sided, whereas the second rejects only excess variability. If fewer than $\lceil0.5|\mathcal V_q|\rceil$ neighbors remain, the implementation keeps that many existing valid neighbors with the smallest $z^{(m)}_{qi}+\max\{z^{(v)}_{qi},0\}$. Invalid or padded slots are never re-enabled and no replacement neighbor is inserted. For retained neighbors available at coordinate $(h,c)$, denoted $\mathcal I_{qhc}$, the consensus-mean proposal is
\begin{equation}
\Delta_{qhc}=\begin{cases}
|\mathcal I_{qhc}|^{-1}\sum_{i\in\mathcal I_{qhc}}\widetilde r_{qihc}, & |\mathcal I_{qhc}|>0,\\
0, & |\mathcal I_{qhc}|=0.
\end{cases}
\label{eq:supp-consensus-proposal}
\end{equation}
The evaluator scores only the issued and observed forecast coordinates represented by $\mathcal H_q$. Its implementation masks are operational details rather than additional paper symbols or data sources, and do not change the mathematical criterion above.

\section{Data Scope, Domain Libraries, and Release Boundary}

The following task manifest fixes the data sources and retrieval boundary. The ten Web subsets are six public GIFT-Eval Web/cloud workload subsets (two Bitbrains \citep{shen2015statistical} and four BizitObs \citep{saha2024bizitobs} logical subsets) and four public LOTSA subsets: Alibaba Cluster Trace 2018 \citep{alibaba2018trace}, Azure virtual-machine traces from 2017 \citep{cortez2017resourcecentral}, Borg Cluster Data 2011 \citep{reiss2011cluster}, and Extended Web Traffic with Missing Values \citep{godahewa2022extended}. The processed Web manifest contains 15,541 univariate series; their fit windows form the Web domain library, from which each query's eligible residual pool is derived, so no test row is evidence. The CGM task uses 109 public ShanghaiT2DM records as targets and four de-identified collections supplied by collaborating data holders as a retrieval-only domain library; ShanghaiT2DM is never evidence. The broader GIFT-Eval diagnostic evaluates 34 subsets satisfying the fixed-window requirements, with each query's eligible residual pool derived from fit windows; it is separate from the primary Web setting.

No raw CGM record, identifier map, physical unit, or source collection name is part of the prepared reproducibility package. Instead, the release-only CGM fixture contains 498 series under release-only identifiers (109 ShanghaiT2DM targets and 389 retrieval-only series). It removes original and parent identifiers, absolute times, units, source names, raw locations, scales, and reversible mappings. Each retained series is independently robust-normalized, clipped to $[-8,8]$, and quantized to three decimals. Thus, the fixture supports functional verification without reproducing the original clinical records or their reported values.

To keep the prepared reproducibility package practical to distribute and run, this fixture uses a fixed 64-step replay grid obtained by subsampling the production stride-32 cache; it also releases frozen point forecasts rather than full predictive distributions. Candidate order and invalid-slot semantics are preserved, but the replay contains fewer windows and uses transformed values. Its rerun metrics can therefore show small numerical differences from the main-paper results, and its point-forecast CRPS is not a replacement for the reported distributional CRPS. The frozen stride-32 ledgers remain the source of the reported results. The public-data setup guide identifies the official acquisition sources and the required local directory contract without distributing source data.

All experiments use a 512-step observed context and a 64-step forecast horizon. A model-specific cache stores frozen forecasts, residuals, representations, candidate lists, correction proposals, and Gate features. The cache is independent for every backbone; residuals are never exchanged across frozen models.

All source records are expanded to univariate series before windowing. A valid cache row contains 512 observed context steps followed by a 64-step forecast horizon, and consecutive candidate starts are separated by 32 steps. Temporal roles are assigned from the future-label boundary before any capacity limit is applied; hence a cap never moves a window from one role into another. For the primary Web task, the fit, model-select, and calibration windows together have a deterministic cap of 4,000 windows per logical subset and 400 per original univariate series. The test side has the same separate limits. The front-90\% allocation is newest-first round-robin within the 70:10:10 role quotas, whereas the test allocation begins with the earliest valid windows after the 90\% boundary. This keeps long series from dominating a logical subset while preserving the chronological role contract.

The broader GIFT-Eval diagnostic uses the same 512/64 window and stride-32 construction, with separate front-90\% and test caps of 2,000 windows per logical subset and 200 per original univariate series. A prepared logical subset enters this diagnostic only when at least one univariate series yields a valid 512+64 window after the observed-mask checks and role construction. Consequently, 34 of the 55 official logical subsets are represented; the remaining 21 are recorded as having no usable window, principally because their series are too short for this window specification. No context or forecast horizon is shortened to admit those subsets.

\section{Temporal Protocol and Leakage Audit}

Target series are divided by future-label boundaries into fit (0--70\%), model-select (70--80\%), calibration (80--90\%), and test (90--100\%) roles. Fit supplies correction and Gate labels; model-select chooses correction checkpoints and tree count; calibration freezes the Gate threshold; test is evaluated only after all such choices are frozen. A window that crosses a boundary may be retrieval-only but never supplies supervision or evaluation for another role.

Web uses the fixed union of the ten logical subsets' fit windows as its domain library; each query's eligible residual pool is derived from that library. Cross-series candidates are allowed, but a candidate from the same parent must finish no later than the query future starts, and a query cannot retrieve itself. CGM uses four de-identified external collections as a retrieval-only domain library. The public ShanghaiT2DM target collection is never an evidence source. The reproducibility materials include parent and source manifests, candidate shortfall and exact-expansion records, and window-level overlap audits. Candidate construction never reads a query future, residual, utility, or test label. A test loader is constructed only after model-selection and threshold decisions are frozen.

\subsection{Audit summary}

The release audit checks source identity, complete processed series, masked $512{+}64$ windows, and the final eligible candidate edges. For Web, the source manifest has no cross-subset exact complete-series overlap; after filtering, all 52,983 valid Chronos-Bolt Base cache windows have distinct complete-window hashes. Test windows have zero exact complete-window or context-only match with the eligible residual pool. Seven future-only content matches are retained as non-leakage audit records: retrieval uses neither the test future nor a test residual. For CGM, ShanghaiT2DM shares no complete processed series and no complete, context-only, or future-only window with the external domain library across all four frozen backbones. The final candidate audits report no target/evidence role violation, self-retrieval, or same-parent temporal-causality violation. Candidate shortfalls remain invalid slots and never trigger a duplicate, test-row, or non-causal substitution.

\subsection{Numerical implementation contract}

The default representation standardizes context by its own location and scale and concatenates that trajectory with the identically standardized frozen forecast. Legal candidates are ranked by squared Euclidean distance. Before aggregation, a neighbor residual is transformed by Equation~\ref{eq:supp-align}; the ratio is clipped to $[1/4,4]$.
\begin{equation}
\widetilde{\mathbf r}_{qi}=\frac{\mathbf r_i}{s_i}\operatorname{clip}\!\left(\frac{s_i}{s_q},\frac14,4\right).
\label{eq:supp-align}
\end{equation}
For each valid neighbor, the mean-centre median-absolute-deviation (MAD) filter uses its available forecast coordinates to compute the mean deviation of the query-scale-aligned residual from the element-wise center and the population standard deviation of that deviation. Across valid neighbors, the two-sided mean-deviation score and one-sided volatility score are normalized by robust dispersion; when that dispersion is zero, the population standard deviation is used instead. A neighbor is retained only if both scores are at most $2.5$. If fewer than 50\% of valid neighbors remain, they are restored in ascending order of mean-deviation score plus non-negative volatility score until the floor is met; invalid or padded slots are never restored and no replacement neighbor is inserted. The proposal averages the retained aligned residuals, with a finite zero-vector fallback.

The Utility-Aware Correction Gate is specified in Appendix~G. It receives a 30-dimensional, inference-visible summary of the consensus-retained evidence and candidate correction; its complete ordered schema, supervision target, fitting protocol, and threshold-freezing rule are part of the documented implementation specification. The default supervision target is positive full-correction MSE utility.

\section{Complete Primary-Result Ledger}

The prepared reproducibility package contains a machine-readable, per-task and per-backbone result ledger. Each record retains the method label, raw-observed aggregation contract, matched frozen-base reference, and MSE/MAE/CRPS percentage improvement. The package verifier recomputes all reported table entries from these records rather than checking a hard-coded final table. The ledger uses only logical task, backbone, and evidence identifiers; it contains no local filesystem paths, raw CGM values, model weights, or checkpoints.

\section{Robustness, Broader GIFT-Eval, and Mechanism Diagnostics}

The reproducibility package contains per-subset records and the complete encoder, neighborhood, Gate-target, GlobalBias, direct learned-regression, cross-backbone Gate-transfer, and four-backbone mechanism results. Their aggregate numerical summaries are reported in Table~\ref{tab:all55-diagnostic}, Tables~\ref{tab:zero-shot-web}--\ref{tab:zero-shot-cgm}, Tables~\ref{tab:appendix-ablation-sensitivity}--\ref{tab:cerr-direct-control}, Table~\ref{tab:appendix-ablation-scope}, Tables~\ref{tab:gate-audit-all}--\ref{tab:error-profile-cgm}, and the Web robustness tables introduced in Section~E.4. The main paper presents the primary results, cross-backbone Gate transfer, and key mechanism comparisons; Table~\ref{tab:cross-backbone-transfer} summarizes the complete transfer records below. The audit computes MSE, MAE, and CRPS per window over issued and observed coordinates and macro-averages them. Cluster bootstrap analyses follow \citet{efron1994bootstrap}; the archive fixes resampling units and seeds.

\subsection{Broader GIFT-Eval diagnostic}

Table~\ref{tab:all55-diagnostic} reports the broader public GIFT-Eval setting, which retains the four temporal roles, an eligible residual pool derived from fit windows, query-augmented retrieval, $K=60$, and held-out testing. Thirty-four of the 55 prepared logical subsets have usable 512+64 windows; 21 are recorded as having no usable window. This setting is reported separately from the primary Web task because its neighborhood structure is intentionally more heterogeneous. The selected diagnostic route uses the raw-z representation, the consensus-mean proposal, and the Utility-Aware Correction Gate with its MSE target.

\begin{table}[t]
\caption{Broader GIFT-Eval diagnostic on 34 subsets satisfying the fixed-window requirements. Values are percentage reductions in raw-observed error; positive is better.}
\label{tab:all55-diagnostic}
\begin{center}
\scriptsize
\setlength{\tabcolsep}{2pt}
\begin{tabular}{@{}lrrrr@{}}
\toprule
Frozen TSFM & MSE gain & MAE gain & CRPS gain & Acceptance rate \\
\midrule
Chronos-2 & +0.168 & +0.052 & -0.089 & 36.15\% \\
Chronos-Bolt Base & +3.533 & +1.409 & +0.955 & 62.26\% \\
TimesFM-2.5 & +0.010 & +0.086 & +0.122 & 13.79\% \\
Toto-2.0-313m & +4.724 & +1.267 & +0.971 & 41.29\% \\
Four-TSFM mean & +2.109 & +0.704 & +0.490 & 38.37\% \\
\bottomrule
\end{tabular}

\end{center}
\end{table}

\subsection{Target sensitivity and cross-TSFM zero-shot Gate transfer}

The complete sensitivity record in Table~\ref{tab:appendix-ablation-sensitivity} varies the representation, removes the frozen forecast from the query, changes $K$, and changes the Gate target while keeping the eligible-pool and chronological contracts fixed. Tables~\ref{tab:zero-shot-web} and~\ref{tab:zero-shot-cgm} separately evaluate whether a frozen Gate fitted on a source TSFM can be applied directly to a target TSFM. The target receives its own frozen forecast, eligible residual-pool overlay, and consensus-mean proposal, but no target fit, model-select, calibration, or threshold update. The source supplies only its frozen trees, feature schema, and threshold. Thus, the study concerns transfer of the output-side Gate rather than transfer of a backbone or hidden-state interface. Each domain contains 12 directed source--target pairs with the diagonal excluded, giving 24 domain--source--target combinations and 72 individual metric comparisons across the two domains. Across all directed pairs in Tables~\ref{tab:zero-shot-web} and~\ref{tab:zero-shot-cgm}, the metric-wise uplifts are $-0.159$, $+0.517$, and $+0.649$ percentage points for MSE, MAE, and CRPS, respectively. Those tables list every source--target--metric cell; the de-identified result ledger retains the corresponding output records. These increments show that the Gate's added value over RACE-TF is metric-dependent, even though the complete transferred pipeline improves every reported comparison against the frozen targets.

Table~\ref{tab:cross-backbone-transfer} summarizes these records using the equal-weight mean of the three percentage improvements, each measured against the matched frozen target. Its diagonal cells use the target-trained RACE results from Table~\ref{tab:main-results}; they are reference values, not cross-backbone transfers. For each source backbone and domain, the Gate uplift column averages the transferred Gate's improvement minus target RACE-TF's improvement over three metrics and three distinct target backbones (nine equally weighted differences). No diagonal value enters this calculation. These differences are measured in percentage points, not as percentage changes relative to RACE-TF's gain. All means are computed from unrounded ledger values and rounded only for display.

All 72 off-diagonal metric results remain positive relative to the matched frozen targets. The domain-wise mean differences between transferred and target-trained RACE are $-0.060/-0.044/-0.063$ percentage points on Web and $-0.008/-0.027/-0.012$ on CGM, in MSE/MAE/CRPS order. Each domain--metric average is therefore within 0.064 percentage points of its target-trained reference; this is a descriptive average, not a claim of equivalence for every pair. Across domains and transfers, the three-metric mean Gate increment over RACE-TF is $+0.336$ percentage points. The complete pipeline also includes the target's residual-evidence proposal, so its positive gains should not be attributed entirely to the transferred Gate.

\begin{table}[t]
\caption{Complete raw-z zero-shot Gate transfer on Web. Gains are relative to the target's matched frozen TSFM; the final column is the equal-weight MSE/MAE/CRPS difference between the transferred Gate and that target's RACE-TF proposal. Positive values are better.}
\label{tab:zero-shot-web}
\begin{center}
\scriptsize
\setlength{\tabcolsep}{3.2pt}
\renewcommand{\arraystretch}{1.04}
\begin{tabular}{@{}llrrrr@{}}
\toprule
Source Gate & Target TSFM & MSE gain (\%) & MAE gain (\%) & CRPS gain (\%) & Mean Gate uplift (pp) \\
\midrule
Chronos-2 & Chronos-Bolt Base & +1.633 & +2.022 & +2.164 & +0.746 \\
Chronos-2 & TimesFM-2.5 & +1.982 & +1.767 & +2.205 & +0.578 \\
Chronos-2 & Toto-2.0-313m & +1.035 & +1.101 & +0.984 & +0.888 \\
Chronos-Bolt Base & Chronos-2 & +1.019 & +0.804 & +0.829 & +0.526 \\
Chronos-Bolt Base & TimesFM-2.5 & +1.189 & +1.091 & +1.339 & -0.200 \\
Chronos-Bolt Base & Toto-2.0-313m & +1.116 & +0.849 & +0.632 & +0.714 \\
TimesFM-2.5 & Chronos-2 & +0.989 & +1.089 & +1.179 & +0.728 \\
TimesFM-2.5 & Chronos-Bolt Base & +1.614 & +2.008 & +2.154 & +0.731 \\
TimesFM-2.5 & Toto-2.0-313m & +1.038 & +1.080 & +0.957 & +0.873 \\
Toto-2.0-313m & Chronos-2 & +0.977 & +1.067 & +1.146 & +0.706 \\
Toto-2.0-313m & Chronos-Bolt Base & +1.638 & +1.998 & +2.132 & +0.729 \\
Toto-2.0-313m & TimesFM-2.5 & +1.975 & +1.764 & +2.198 & +0.573 \\
\bottomrule
\end{tabular}

\end{center}
\end{table}

\begin{table}[t]
\caption{Complete raw-z zero-shot Gate transfer on CGM. Gains are relative to the target's matched frozen TSFM; the final column is the equal-weight MSE/MAE/CRPS difference between the transferred Gate and that target's RACE-TF proposal. Positive values are better.}
\label{tab:zero-shot-cgm}
\begin{center}
\scriptsize
\setlength{\tabcolsep}{3.2pt}
\renewcommand{\arraystretch}{1.04}
\begin{tabular}{@{}llrrrr@{}}
\toprule
Source Gate & Target TSFM & MSE gain (\%) & MAE gain (\%) & CRPS gain (\%) & Mean Gate uplift (pp) \\
\midrule
Chronos-2 & Chronos-Bolt Base & +22.801 & +12.895 & +10.672 & +0.129 \\
Chronos-2 & TimesFM-2.5 & +19.862 & +9.879 & +9.244 & +0.117 \\
Chronos-2 & Toto-2.0-313m & +22.235 & +11.364 & +10.426 & +0.048 \\
Chronos-Bolt Base & Chronos-2 & +20.708 & +10.609 & +9.895 & +0.027 \\
Chronos-Bolt Base & TimesFM-2.5 & +19.684 & +9.818 & +9.177 & +0.015 \\
Chronos-Bolt Base & Toto-2.0-313m & +22.228 & +11.388 & +10.418 & +0.051 \\
TimesFM-2.5 & Chronos-2 & +20.448 & +10.611 & +9.843 & -0.077 \\
TimesFM-2.5 & Chronos-Bolt Base & +22.751 & +12.921 & +10.663 & +0.118 \\
TimesFM-2.5 & Toto-2.0-313m & +21.886 & +11.245 & +10.277 & -0.158 \\
Toto-2.0-313m & Chronos-2 & +20.712 & +10.573 & +9.877 & +0.010 \\
Toto-2.0-313m & Chronos-Bolt Base & +22.735 & +12.914 & +10.665 & +0.111 \\
Toto-2.0-313m & TimesFM-2.5 & +19.774 & +9.857 & +9.216 & +0.071 \\
\bottomrule
\end{tabular}

\end{center}
\end{table}

\subsection{Complete four-backbone ablation and mechanism records}

Tables~\ref{tab:appendix-ablation-sensitivity}--\ref{tab:appendix-ablation-scope} materialize the ablations that are intentionally outside the body mechanism table. Each entry is the unweighted four-backbone average percentage improvement relative to its matched frozen base, ordered as MSE / MAE / CRPS. Positive values mean lower error. The fixed body configuration uses the raw-z representation, an eligible residual pool drawn from the domain library, history plus frozen forecast in the query, residual consensus, $K=60$, and the Utility-Aware Correction Gate with its MSE target; every row below names its deviation from that configuration.

\begin{table}[t]
\caption{Complete primary mechanism ablations. These controls are diagnostic variants, not additional RACE variants.}
\label{tab:appendix-ablation-sensitivity}
\begin{center}
\scriptsize
\setlength{\tabcolsep}{2pt}
\renewcommand{\arraystretch}{1.05}
\begin{tabularx}{\linewidth}{@{}>{\raggedright\arraybackslash}p{.17\linewidth}>{\raggedright\arraybackslash}Xcc@{}}
\toprule
Group & Configuration and changed factor & Web (\%) & CGM (\%) \\
 &  & MSE / MAE / CRPS & MSE / MAE / CRPS \\
\midrule
Deployed route & RACE: raw-z representation, Consensus Mean, $K=60$, Gate with MSE target & \metric{+1.410 / +1.431 / +1.557} & \metric{+21.327 / +11.200 / +10.043} \\
Encoder sensitivity & HybridMath representation replaces raw-z & \metric{+1.807 / +1.256 / +1.337} & \metric{+20.637 / +10.725 / +9.300} \\
Query control & Chronos-T5 final-token representation; history-only query & \metric{-0.071 / -0.312 / -0.406} & \metric{+6.330 / +2.145 / +1.815} \\
Query control & History only; frozen forecast removed from raw-z representation & \metric{+1.214 / +1.209 / +1.336} & \metric{+16.484 / +8.943 / +8.021} \\
Neighborhood & $K=10$ replaces $K=60$ & \metric{+1.486 / +1.035 / +1.052} & \metric{+14.336 / +6.549 / +5.735} \\
Neighborhood & $K=30$ replaces $K=60$ & \metric{+1.528 / +1.366 / +1.471} & \metric{+20.088 / +10.109 / +9.128} \\
Evidence control & Mean of all retrieved residuals; no consensus & \metric{+2.247 / +0.984 / +0.917} & \metric{+21.716 / +10.154 / +9.086} \\
Selective transfer & RACE-TF: Consensus Mean applied to every query & \metric{+1.646 / +0.443 / +0.243} & \metric{+21.341 / +11.083 / +9.983} \\
Direct learned control & CERR: learned residual regressor on the same consensus-retained evidence; applied to every query & \metric{+1.584 / -6.467 / -9.042} & \metric{+26.629 / +11.887 / +9.955} \\
Gate target & Symmetric mean absolute percentage error (sMAPE) target replaces the default MSE target & \metric{+1.385 / +1.383 / +1.487} & \metric{+20.536 / +10.941 / +9.778} \\
Gate target & MAE target replaces the default MSE target & \metric{+1.346 / +1.333 / +1.463} & \metric{+20.316 / +10.859 / +9.694} \\
Gate target & CRPS target replaces the default MSE target & \metric{+1.348 / +1.341 / +1.520} & \metric{+20.527 / +10.980 / +9.798} \\
Retrieval-free control & Horizon-wise GlobalBias, without Gate & \metric{-13.563 / -21.914 / -27.024} & \metric{+0.543 / -1.706 / -1.341} \\
\bottomrule
\end{tabularx}

\end{center}
\end{table}

\subsubsection{Direct learned residual-regression control}

The Consensus-Evidence Residual Regressor (CERR) is a matched diagnostic control for separating the value of a learned residual mapping from the value of utility-aware selection. It preserves the deployed raw-z representation, the eligible residual pool derived from the domain library, exact $K=60$ retrieval, query-scale residual alignment, and the identical mask produced by mean-centred, MAD-based residual consensus used by RACE-TF and RACE. It therefore receives only evidence available after Residual Evidence Construction: the most recent 128 raw-z context values, the 64-step frozen forecast standardized with the same context statistics, every retained aligned neighbor-residual trajectory, standardized retrieval scores, clipped scale ratios, the valid-neighbor mask, and the exact consensus-mean proposal. A shared neighbor network, mask-aware pooling, and a small multilayer output head map these inputs to one bounded residual correction. Invalid slots and slots rejected by residual consensus have no contribution, and an empty retained set deterministically emits the zero correction. CERR has no Gate, probability, or test-time threshold: it applies its learned correction to every query.

CERR is fitted only on the fit role by direct MSE minimization (AdamW, learning rate $3\times10^{-4}$, weight decay $10^{-4}$, batch size 64, at most 50 epochs). The 70--80\% model-select role chooses the epoch after at least eight epochs with patience five. The 80--90\% calibration role is not used because this control has no threshold, and the 90--100\% test role is first read only after model selection. Thus, CERR does not use future labels at inference and does not use test labels for fitting or design selection.

Table~\ref{tab:cerr-direct-control} gives the complete matched comparison. On CGM, the four-backbone means favor CERR for MSE and MAE and RACE for CRPS among these three output-side controls. On Web, CERR obtains a positive MSE reduction but substantially harms MAE and CRPS, including a large degradation for Toto-2.0-313m. RACE is not uniformly best on every single metric; rather, the contrast shows that directly learning a correction from the same evidence can recover squared-error gains in a favorable task while failing to provide stable multi-metric transfer. The Utility-Aware Correction Gate is therefore supported as a harm-control mechanism, not as a claim that selection maximizes every metric for every backbone.

\begin{table}[t]
\caption{Matched direct residual-regression control. All rows use the same raw-z context-plus-frozen-forecast representation, eligible residual pool derived from the domain library, $K=60$, evidence retained by mean-centred MAD-based residual consensus, raw-observed evaluator, test windows, and frozen TSFM. CERR replaces only the consensus-mean output interface with a learned residual regressor applied to every query; it has no Gate. Values are percentage reductions in error relative to the matched frozen base, ordered as MSE / MAE / CRPS; positive is better. Four-TSFM means are unweighted.}
\label{tab:cerr-direct-control}
\begin{center}
\scriptsize
\setlength{\tabcolsep}{1.5pt}
\renewcommand{\arraystretch}{1.05}
\begin{tabular}{@{}llccc@{}}
\toprule
Task & Frozen TSFM & RACE-TF (all queries) & CERR (direct learned) & RACE (Gate-selected) \\
 & & MSE / MAE / CRPS & MSE / MAE / CRPS & MSE / MAE / CRPS \\
\midrule
Web & Chronos-2 & \metric{+1.159 / +0.099 / -0.186} & \metric{+1.445 / -8.179 / -12.516} & \metric{+0.982 / +1.091 / +1.182} \\
Web & Chronos-Bolt Base & \metric{+1.782 / +0.954 / +0.846} & \metric{+4.290 / -3.513 / -4.963} & \metric{+1.647 / +1.838 / +1.945} \\
Web & TimesFM-2.5 & \metric{+2.342 / +0.845 / +1.033} & \metric{+2.826 / -4.014 / -4.782} & \metric{+1.986 / +1.748 / +2.187} \\
Web & Toto-2.0-313m & \metric{+1.301 / -0.126 / -0.719} & \metric{-2.226 / -10.163 / -13.909} & \metric{+1.025 / +1.047 / +0.912} \\
\textbf{Web} & \textbf{Four-TSFM mean} & \metric{+1.646 / +0.443 / +0.243} & \metric{+1.584 / -6.467 / -9.042} & \metric{+1.410 / +1.431 / +1.557} \\
\midrule
CGM & Chronos-2 & \metric{+20.752 / +10.531 / +9.849} & \metric{+22.886 / +9.697 / +8.208} & \metric{+20.814 / +10.677 / +9.939} \\
CGM & Chronos-Bolt Base & \metric{+22.637 / +12.769 / +10.575} & \metric{+25.920 / +13.114 / +10.333} & \metric{+22.773 / +12.928 / +10.688} \\
CGM & TimesFM-2.5 & \metric{+19.735 / +9.752 / +9.147} & \metric{+28.287 / +12.304 / +10.711} & \metric{+19.686 / +9.927 / +9.233} \\
CGM & Toto-2.0-313m & \metric{+22.239 / +11.281 / +10.362} & \metric{+29.421 / +12.431 / +10.568} & \metric{+22.032 / +11.268 / +10.311} \\
\textbf{CGM} & \textbf{Four-TSFM mean} & \metric{+21.341 / +11.083 / +9.983} & \metric{+26.629 / +11.887 / +9.955} & \metric{+21.327 / +11.200 / +10.043} \\
\bottomrule
\end{tabular}

\end{center}
\end{table}

Table~\ref{tab:appendix-ablation-scope} isolates restricted-pool controls under the same raw-observed evaluator. These rows isolate the effect of restricting the eligible residual pool relative to the default full-domain construction.

\begin{table}[t]
\caption{Web-only retrieval-scope diagnostics and restricted-pool controls. CGM is omitted because its external domain library and held-out target split differ from the Web protocol. These controls isolate the effect of restricting the eligible residual pool relative to the default full-domain construction. The default Gate uses MSE utility; the sMAPE-targeted Gate in Table~\ref{tab:appendix-ablation-sensitivity} is a separate sensitivity control.}
\label{tab:appendix-ablation-scope}
\begin{center}
\footnotesize
\setlength{\tabcolsep}{3pt}
\renewcommand{\arraystretch}{1.05}
\begin{tabularx}{\linewidth}{@{}>{\raggedright\arraybackslash}p{.18\linewidth}>{\raggedright\arraybackslash}Xc@{}}
\toprule
Group & Configuration and changed factor & Web (\%) \\
 &  & MSE / MAE / CRPS \\
\midrule
Scope diagnostic & RACE with raw-z representation; same logical-subset eligible residual pool only & \metric{-0.277 / +0.378 / +0.288} \\
Scope diagnostic & RACE with HybridMath representation; same logical-subset eligible residual pool only & \metric{+0.106 / +0.501 / +0.523} \\
Scope diagnostic & RACE with raw-z representation; hard Top-3 subset-card eligible residual pool & \metric{-0.769 / +0.016 / -0.198} \\
Scope diagnostic & RACE with HybridMath representation; hard Top-3 subset-card eligible residual pool & \metric{+0.938 / +1.004 / +1.056} \\
Unconditional transfer & Same-subset mean, Web only; no consensus or Gate & \metric{-39.280 / -26.983 / -35.841} \\
Gated transfer & Same-subset mean, Web only; Gate with default MSE target & \metric{-1.662 / -1.472 / -2.013} \\
\bottomrule
\end{tabularx}

\end{center}
\end{table}

Table~\ref{tab:gate-audit-all} expands the Chronos-Bolt Base illustration in Figure~\ref{fig:selective-audit}\protect\subref{fig:action-audit} by recording each backbone's frozen threshold, acceptance rate, and post-hoc MSE utility diagnostic. Tables~\ref{tab:error-profile-web} and~\ref{tab:error-profile-cgm} likewise expand Figure~\ref{fig:selective-audit}\protect\subref{fig:error-stratified} with all five within-dataset frozen-base-MSE bins for every backbone and task. These test-label summaries are generated only after all model states and thresholds are frozen.

\begin{table}[t]
\caption{Complete four-backbone Figure~\ref{fig:selective-audit}\protect\subref{fig:action-audit} Gate audit. RACE-TF beneficial and Benefit among accepted are post-hoc fractions of test windows whose RACE-TF proposal lowers per-window MSE. AP is average precision for this beneficial-correction label; Spearman is the rank correlation between Gate probability and per-window RACE-TF MSE percentage gain. All diagnostics are computed after thresholds are frozen.}
\label{tab:gate-audit-all}
\begin{center}
\scriptsize
\setlength{\tabcolsep}{2.7pt}
\renewcommand{\arraystretch}{1.04}
\begin{tabular}{@{}llrrrrrr@{}}
\toprule
Task & Frozen TSFM & $\tau$ & \shortstack{RACE-TF\\beneficial (\%)} & \shortstack{Acceptance\\rate (\%)} & \shortstack{Benefit among\\accepted (\%)} & AP & Spearman \\
\midrule
CGM & Chronos-2 & 0.514 & 68.09 & 95.25 & 69.05 & 0.773 & 0.336 \\
CGM & Chronos-Bolt Base & 0.601 & 69.91 & 95.74 & 71.63 & 0.780 & 0.400 \\
CGM & TimesFM-2.5 & 0.581 & 66.35 & 90.55 & 68.32 & 0.744 & 0.306 \\
CGM & Toto-2.0-313m & 0.644 & 69.42 & 90.67 & 70.96 & 0.787 & 0.368 \\
Web & Chronos-2 & 0.453 & 37.85 & 52.02 & 52.69 & 0.548 & 0.520 \\
Web & Chronos-Bolt Base & 0.564 & 40.23 & 51.75 & 53.48 & 0.529 & 0.421 \\
Web & TimesFM-2.5 & 0.503 & 39.12 & 46.97 & 56.95 & 0.587 & 0.555 \\
Web & Toto-2.0-313m & 0.512 & 38.88 & 39.32 & 54.51 & 0.513 & 0.400 \\
\bottomrule
\end{tabular}

\end{center}
\end{table}

\begin{table}[t]
\caption{Complete four-backbone Figure~\ref{fig:selective-audit}\protect\subref{fig:error-stratified} error-stratified audit on Web. Error bins are within-dataset equal-count quintiles of frozen-base MSE (1=lowest, 5=highest). MSE reductions are dataset-macro averages after normalizing each dataset by its mean frozen-base MSE. Acceptance rate and Benefit among accepted are post-hoc test diagnostics; $n$ is the number of windows in the bin.}
\label{tab:error-profile-web}
\begin{center}
\scriptsize
\setlength{\tabcolsep}{3.1pt}
\renewcommand{\arraystretch}{1.02}
\begin{tabular}{@{}llrrrrr@{}}
\toprule
Frozen TSFM & Error bin & \shortstack{RACE MSE\\reduction (\%)} & \shortstack{RACE-TF MSE\\reduction (\%)} & \shortstack{Acceptance\\rate (\%)} & \shortstack{Benefit among\\accepted (\%)} & $n$ \\
\midrule
Chronos-2 & 1 & -0.02 & -30.99 & 34.48 & 36.34 & 4820 \\
Chronos-2 & 2 & -0.12 & -60.11 & 42.13 & 50.89 & 4814 \\
Chronos-2 & 3 & -0.52 & -51.59 & 49.74 & 53.86 & 4815 \\
Chronos-2 & 4 & +0.07 & -26.84 & 67.30 & 57.13 & 4814 \\
Chronos-2 & 5 & +60.92 & +42.34 & 66.48 & 56.96 & 4812 \\
Chronos-Bolt Base & 1 & -0.50 & -35.68 & 39.98 & 34.61 & 4820 \\
Chronos-Bolt Base & 2 & -2.18 & -89.26 & 47.30 & 53.67 & 4814 \\
Chronos-Bolt Base & 3 & -2.11 & -53.57 & 48.16 & 58.09 & 4815 \\
Chronos-Bolt Base & 4 & -1.06 & -21.65 & 57.25 & 62.63 & 4814 \\
Chronos-Bolt Base & 5 & +55.83 & +33.94 & 66.11 & 53.51 & 4812 \\
TimesFM-2.5 & 1 & +0.01 & -34.49 & 29.38 & 46.40 & 4820 \\
TimesFM-2.5 & 2 & -0.06 & -69.66 & 38.82 & 55.48 & 4814 \\
TimesFM-2.5 & 3 & -0.19 & -65.29 & 44.13 & 58.45 & 4815 \\
TimesFM-2.5 & 4 & +0.59 & -24.85 & 61.57 & 60.32 & 4814 \\
TimesFM-2.5 & 5 & +55.10 & +28.67 & 60.99 & 58.47 & 4812 \\
Toto-2.0-313m & 1 & -0.52 & -30.78 & 30.35 & 34.52 & 4820 \\
Toto-2.0-313m & 2 & -0.57 & -65.90 & 33.22 & 51.84 & 4814 \\
Toto-2.0-313m & 3 & -1.68 & -44.67 & 33.21 & 54.97 & 4815 \\
Toto-2.0-313m & 4 & -0.36 & -17.71 & 44.08 & 61.92 & 4814 \\
Toto-2.0-313m & 5 & +58.35 & +48.48 & 55.78 & 60.84 & 4812 \\
\bottomrule
\end{tabular}

\end{center}
\end{table}

\begin{table}[t]
\caption{Complete four-backbone Figure~\ref{fig:selective-audit}\protect\subref{fig:error-stratified} error-stratified audit on CGM. Error bins are within-dataset equal-count quintiles of frozen-base MSE (1=lowest, 5=highest). MSE reductions are dataset-macro averages after normalizing each dataset by its mean frozen-base MSE. Acceptance rate and Benefit among accepted are post-hoc test diagnostics; $n$ is the number of windows in the bin.}
\label{tab:error-profile-cgm}
\begin{center}
\scriptsize
\setlength{\tabcolsep}{3.1pt}
\renewcommand{\arraystretch}{1.02}
\begin{tabular}{@{}llrrrrr@{}}
\toprule
Frozen TSFM & Error bin & \shortstack{RACE MSE\\reduction (\%)} & \shortstack{RACE-TF MSE\\reduction (\%)} & \shortstack{Acceptance\\rate (\%)} & \shortstack{Benefit among\\accepted (\%)} & $n$ \\
\midrule
Chronos-2 & 1 & -1.74 & -1.77 & 96.36 & 47.80 & 165 \\
Chronos-2 & 2 & +1.21 & +1.02 & 94.51 & 65.16 & 164 \\
Chronos-2 & 3 & +7.20 & +6.91 & 92.07 & 69.54 & 164 \\
Chronos-2 & 4 & +22.32 & +21.96 & 98.17 & 80.75 & 164 \\
Chronos-2 & 5 & +75.22 & +75.76 & 95.12 & 82.05 & 164 \\
Chronos-Bolt Base & 1 & -4.08 & -4.16 & 95.76 & 43.04 & 165 \\
Chronos-Bolt Base & 2 & +3.32 & +3.27 & 95.73 & 69.43 & 164 \\
Chronos-Bolt Base & 3 & +12.23 & +12.13 & 96.95 & 78.62 & 164 \\
Chronos-Bolt Base & 4 & +22.09 & +21.75 & 95.12 & 78.85 & 164 \\
Chronos-Bolt Base & 5 & +80.47 & +80.35 & 95.12 & 88.46 & 164 \\
TimesFM-2.5 & 1 & -3.47 & -3.62 & 91.76 & 42.95 & 170 \\
TimesFM-2.5 & 2 & -0.94 & -1.11 & 89.35 & 62.91 & 169 \\
TimesFM-2.5 & 3 & +8.05 & +7.20 & 88.82 & 74.17 & 170 \\
TimesFM-2.5 & 4 & +19.22 & +19.11 & 92.31 & 77.56 & 169 \\
TimesFM-2.5 & 5 & +75.78 & +77.31 & 90.53 & 84.31 & 169 \\
Toto-2.0-313m & 1 & -3.18 & -3.33 & 94.12 & 47.50 & 170 \\
Toto-2.0-313m & 2 & +2.58 & +2.67 & 91.72 & 68.39 & 169 \\
Toto-2.0-313m & 3 & +9.19 & +9.13 & 87.06 & 75.00 & 170 \\
Toto-2.0-313m & 4 & +20.89 & +21.23 & 89.35 & 78.81 & 169 \\
Toto-2.0-313m & 5 & +80.92 & +81.72 & 91.12 & 86.36 & 169 \\
\bottomrule
\end{tabular}

\end{center}
\end{table}

\subsubsection{Graphical extensions of Figure~\ref{fig:selective-audit} for the remaining backbones}

\begin{figure}[t]
\centering
\begin{minipage}[t]{0.475\textwidth}
\centering
\includegraphics[width=\linewidth]{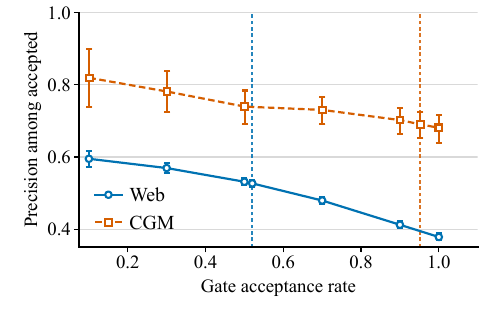}\\[-0.8ex]
\small\textbf{(a) Figure~\ref{fig:selective-audit}\protect\subref{fig:action-audit} extension: Chronos-2}
\end{minipage}\hfill
\begin{minipage}[t]{0.475\textwidth}
\centering
\includegraphics[width=\linewidth]{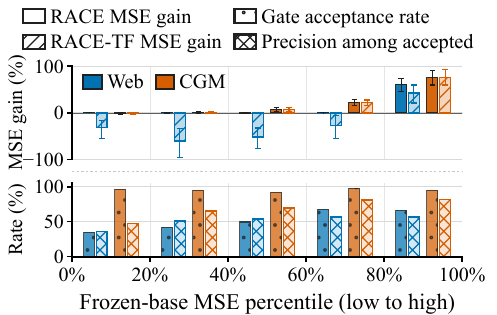}\\[-0.8ex]
\small\textbf{(b) Figure~\ref{fig:selective-audit}\protect\subref{fig:error-stratified} extension: Chronos-2}
\end{minipage}
\caption{Selective residual-transfer audit extension: Chronos-2. The left and right panels reproduce the respective body diagnostic procedures without model, threshold, or test-time selection changes.}
\label{fig:appendix-d2-chronos2}
\end{figure}

\begin{figure}[t]
\centering
\begin{minipage}[t]{0.475\textwidth}
\centering
\includegraphics[width=\linewidth]{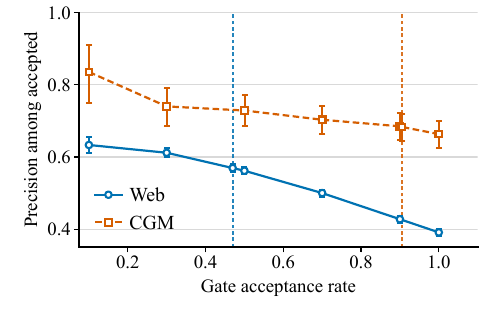}\\[-0.8ex]
\small\textbf{(a) Figure~\ref{fig:selective-audit}\protect\subref{fig:action-audit} extension: TimesFM-2.5}
\end{minipage}\hfill
\begin{minipage}[t]{0.475\textwidth}
\centering
\includegraphics[width=\linewidth]{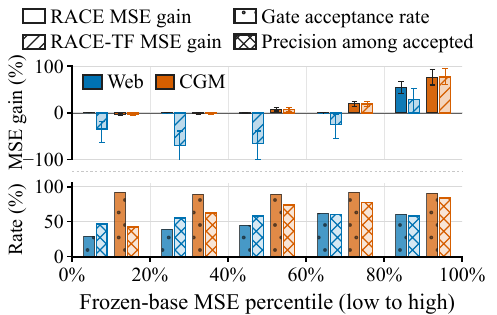}\\[-0.8ex]
\small\textbf{(b) Figure~\ref{fig:selective-audit}\protect\subref{fig:error-stratified} extension: TimesFM-2.5}
\end{minipage}
\caption{Selective residual-transfer audit extension: TimesFM-2.5. Definitions, threshold provenance, and bootstrap units are identical to Figure~\ref{fig:selective-audit}.}
\label{fig:appendix-d2-timesfm}
\end{figure}

\begin{figure}[t]
\centering
\begin{minipage}[t]{0.475\textwidth}
\centering
\includegraphics[width=\linewidth]{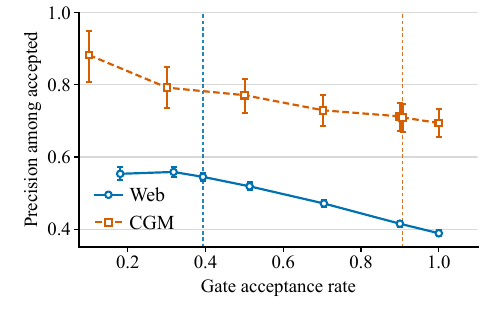}\\[-0.8ex]
\small\textbf{(a) Figure~\ref{fig:selective-audit}\protect\subref{fig:action-audit} extension: Toto-2.0-313m}
\end{minipage}\hfill
\begin{minipage}[t]{0.475\textwidth}
\centering
\includegraphics[width=\linewidth]{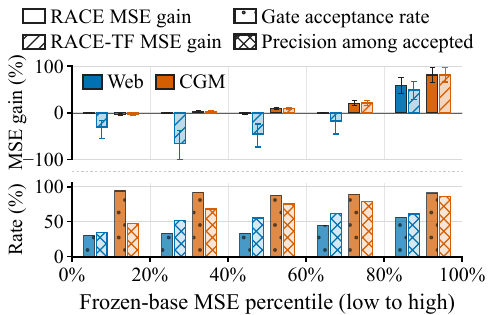}\\[-0.8ex]
\small\textbf{(b) Figure~\ref{fig:selective-audit}\protect\subref{fig:error-stratified} extension: Toto-2.0-313m}
\end{minipage}
\caption{Selective residual-transfer audit extension: Toto-2.0-313m. The panels retain the same task encodings and outcome definitions as all other supplementary extensions.}
\label{fig:appendix-d2-toto}
\end{figure}

Figures~\ref{fig:appendix-d2-chronos2}, \ref{fig:appendix-d2-timesfm}, and~\ref{fig:appendix-d2-toto} render the same two diagnostic procedures used in body Figure~\ref{fig:selective-audit} for Chronos-2, TimesFM-2.5, and Toto-2.0-313m. The Chronos-Bolt Base pair is already displayed in the body and is deliberately not duplicated here. For each remaining backbone, the left panel reports post-hoc RACE-TF MSE precision as the Gate acceptance rate varies; the task-colored dashed line is the acceptance rate induced by the saved calibration-frozen threshold. The right panel partitions each dataset into five frozen-base-MSE percentile ranges and reports RACE and RACE-TF MSE gains, plus acceptance rate and precision. Each panel is regenerated from the frozen per-sample replay ledger using the same stable binning and 2,000 target-parent bootstrap resamples as the body figures. These diagnostic test labels are never used to refit a Gate, choose a threshold, or alter inference.

\subsection{Web dataset-level robustness of RACE}

The Web result in the body follows the pre-specified raw-observed, test-window macro average. Table~\ref{tab:web-subset-win-loss} provides a compact, complementary coverage audit across the ten logical Web subsets and four frozen TSFMs. Each cell first aggregates raw-observed test-window errors within one subset and one TSFM, then computes the matched-base percentage improvement. A positive improvement is counted as a win; a non-positive improvement is counted as a loss. Thus, every subset--TSFM cell is counted exactly once.

The resulting W/L counts show the effect of utility-aware selection at the subset level. For MSE, RACE records 26 wins and 14 losses, compared with 21 and 19 for RACE-TF. For MAE and CRPS, RACE increases wins from 10 to 17 and from 11 to 17, while reducing losses from 30 to 23 and from 29 to 23, respectively. The counts therefore show that the Gate reduces the cross-subset harm of unconditional transfer, while not implying that RACE improves every individual subset--TSFM cell. All quantities are computed only after models and calibration thresholds are frozen; no test label was used to alter a forecast, fit a Gate, select a threshold, or choose the deployed route.

\begin{table}[t]
\caption{Web subset--backbone win/loss summary. For each metric, W and L count the 40 logical-subset--frozen-TSFM cells with, respectively, positive and non-positive matched-base percentage improvement. Thus W+L equals 40 for every method and metric. These frozen post-hoc counts do not select a route, model state, or Gate threshold.}
\label{tab:web-subset-win-loss}
\begin{center}
\scriptsize
\setlength{\tabcolsep}{5pt}
\renewcommand{\arraystretch}{1.05}
\begin{tabular}{@{}lccc@{}}
\toprule
Method & MSE W/L & MAE W/L & CRPS W/L \\
\midrule
RACE-TF & 21/19 & 10/30 & 11/29 \\
RACE & 26/14 & 17/23 & 17/23 \\
RAF & 13/27 & 4/36 & 8/32 \\
\bottomrule
\end{tabular}

\end{center}
\end{table}

\section{Comparator Implementation and Evaluation}

RAF follows the input-augmentation procedure of \citet{tire2024raf} across all four frozen backbones. It retrieves one historical continuation using the sum of token-wise Euclidean distances between native frozen context representations, independently normalizes the retrieved 512-step context plus 64-step future and the 512-step query context, aligns their endpoints, and reforecasts from the concatenated 1088-step input. Chronos-2 uses context embedding tokens, Chronos-Bolt Base uses encoder patch states, TimesFM-2.5 uses decoder prefill states, and Toto-2.0-313m uses context transformer states. No fitting, mixing coefficient, or test-label selection is used. Table~\ref{tab:main-results} reports this input-augmented comparator alongside the residual-correction methods.

Web uses the fixed fit-window candidate pool with same-parent causal filtering, and CGM uses external retrieval-only memory. Candidates require a fully observed 64-step continuation. Each method is evaluated against its backbone's immutable base ledger on issued and observed forecast coordinates. Web windows, labels, masks, and roles match across all four backbones. CGM retains the existing per-backbone samples: 821 windows for Chronos-2 and Chronos-Bolt Base, and 847 for TimesFM-2.5 and Toto-2.0-313m. Thus, Table~\ref{tab:main-results} compares methods within each matched backbone, not four backbones on an identical CGM sample set.

Point and quantile adapters are held fixed relative to the matching base forecasts. In particular, TimesFM-2.5 retains the original nine-channel mapping obtained by removing channel index 5 from its ten native output channels. Its CRPS therefore follows the project's existing evaluation convention, rather than a newly calibrated quantile mapping. The input-augmented RAF records are checked against per-window errors, causal candidate audits, and base-ledger hashes.

Native TS-RAG is evaluated only with its supported Chronos-Bolt Base interface \citep{ning2025tsrag}. Its fine-tuned condition freezes Chronos-Bolt Base and Chronos-T5 and updates only the release-provided retrieval-augmentation module using AdamW \citep{loshchilov2019adamw}, learning rate $10^{-4}$, weight decay 0.01, batch size 8, and at most eight epochs; model-select chooses the epoch and calibration remains held out.

CERR is an internal matched control rather than an external retrieval baseline. It reuses the complete RACE-TF evidence pipeline and changes only the final output interface: a compact supervised residual regressor replaces the parameter-free consensus-mean proposal, while the Utility-Aware Correction Gate is absent. Its architecture and chronology are specified in Appendix~E.3 so that the comparison isolates direct learned residual transfer from selective residual transfer.

The backbone manifest records the exact Chronos-2, Chronos-Bolt Base, TimesFM-2.5, and Toto-2.0-313m versions. The latter is pinned to the official Datadog/Toto-2.0-313m model card \citep{toto2026modelcard} and to the Toto~2 paper \citep{khwaja2026toto}, including revision and access date in the prepared reproducibility package. The prepared code package provides the corresponding concrete inference-environment configuration and checks its declared runtime prerequisites before execution.

\section{Utility-Aware Correction Gate Contract}

This appendix fixes the input, output, and fitting contract of the Utility-Aware Correction Gate. The ordered feature schema is serialized and hashed with each fitted model; inference fails if feature membership or order differs from the fitted schema.

\begin{table}[t]
\caption{Exact 30-dimensional, inference-visible schema for the Utility-Aware Correction Gate. The default target is positive full-correction MSE utility.}
\label{tab:gate-schema}
\begin{center}
\footnotesize
\setlength{\tabcolsep}{3pt}
\renewcommand{\arraystretch}{1.08}
\begin{tabular}{@{}p{0.20\textwidth}p{0.47\textwidth}p{0.25\textwidth}@{}}
\toprule
Feature block (dimension) & Exact construction & Decision information \\
\midrule
Evidence availability (2) & Consensus-retained valid-neighbor count and valid-neighbor fraction. & Amount of legal, consensus-retained evidence. \\
Aligned residual evidence (9) & Mean and standard deviation of values and of their absolute values, median, and root-mean-square over valid aligned residual coordinates; mean and standard deviation of within-neighbor residual standard deviations; and $1/(1+\operatorname{mean}\text{ within-neighbor standard deviation})$. & Residual level, magnitude, variability, and temporal coherence. \\
Retrieval similarity (8) & Mean, standard deviation, minimum, maximum, 10th/50th/90th percentiles, and Gini coefficient of $1/(1+\lVert\mathbf e_i-\mathbf e_q\rVert_2)$ over valid neighbors. & Closeness and concentration of retrieved support. \\
Frozen forecast shape (2) & Standard deviation and mean absolute first difference of the query-scale-normalized frozen forecast, $\operatorname{clip}(\hat{\mathbf y}_q/s_q,-50,50)$. & Volatility of the backbone forecast. \\
Observed context shape (2) & Standard deviation and mean absolute first difference of observed $\mathbf c_q/s_q$. & Recent target dynamics in comparable units. \\
Consensus proposal (3) & Mean absolute value, standard deviation, and maximum absolute value of $\boldsymbol\Delta_q$ on $\mathcal H_q$. & Size and temporal variation of the proposed correction. \\
Proposal weights (3) & Maximum, Shannon entropy, and effective count $1/\max(\sum_i w_i^2,10^{-12})$, clipped at $10^4$. & Retained evidence support. \\
Direction consistency (1) & Indicator that the product of the mean aligned residual and mean $\boldsymbol\Delta_q$ is non-negative. & Whether the proposed direction agrees with retained evidence. \\
\bottomrule
\end{tabular}

\end{center}
\end{table}

Within each row of Table~\ref{tab:gate-schema}, fields are serialized in the stated order. The prepared reproducibility package exposes the corresponding code-style field names and the schema hash; inference fails if that ordered sequence differs from the fitted model metadata. This preserves an exact executable contract without placing an unbreakable 30-name code string in the typeset supplement.

All neighbor-derived quantities use the common consensus-valid mask, neighbor forecast mask, and $\mathcal H_q$; context quantities use the context-observation mask. For the default consensus mean, $w_i=1/n$ on the $n>0$ retained neighbors with at least one valid forecast coordinate, and zero elsewhere; with no evidence, all weights and the correction are zero, while the stabilized effective-count feature is $10^4$ after denominator flooring and feature clipping. Direction agreement tests whether the product of the pooled residual mean and coordinate-averaged proposal is non-negative: it is one under identical complete support (including zero means), but can differ under unequal coordinate masks. The schema omits absolute level, TSFM identity, embedding coordinates, per-horizon trajectories, explicit consensus diagnostics, and every query future, realized residual, utility, or test label. It therefore captures the inference-visible consequences of residual evidence construction and its resulting proposal without using an outcome unavailable at deployment.

For every development window, $b_q=1$ exactly when applying the complete consensus proposal reduces valid-horizon MSE relative to the frozen forecast. We fit a binary logistic XGBoost classifier with maximum tree depth 3, learning rate 0.02, at most 300 trees, row-subsampling rate 0.6, feature-subsampling rate 0.6, minimum child weight 50, L2 regularization 10, L1 regularization 0.5, and seed 42. A provisional fit-window model uses model-select PR-AUC with 30-round early stopping to choose the tree count. The classifier is then refitted on fit plus model-select windows with hierarchy-balanced weights across source family, dataset, parent series, and windows. On calibration, $\tau$ is selected from 50 score quantiles between 5\% and 95\% to maximize total accepted MSE utility. The deployed output is $a_q=\mathbf{1}[p_q\geq\tau]$; the test loader is created only after this procedure completes.

\section{Figure Provenance and Case Selection}

\noindent\textbf{Figure 1.} Illustrative Web and CGM windows follow a pre-recorded case-selection rule; they are not selected by observed test improvement. Target, residual-evidence, Gate, and neighbor roles are rendered with vector and grayscale checks.

\noindent\textbf{Figure 2.} This is a conceptual pipeline diagram and contains no result-dependent numerical quantity.

\noindent\textbf{Figure 3.} The plot combines four-backbone calibration-time residual-coherence statistics with a post-hoc test-utility column. Its uncertainty intervals use target-parent bootstrap resampling with 2,000 draws and seed 42.

\noindent\textbf{Figure~\ref{fig:selective-audit}\protect\subref{fig:action-audit}.} Chronos-Bolt Base is displayed in the body for readability under a pre-existing route criterion. Figures~\ref{fig:appendix-d2-chronos2}--\ref{fig:appendix-d2-toto} provide the corresponding plots for the remaining three backbones, and Table~\ref{tab:gate-audit-all} provides the complete four-backbone Gate audit.

\noindent\textbf{Figure~\ref{fig:selective-audit}\protect\subref{fig:error-stratified}.} The panels use five within-dataset frozen-base-MSE percentile bins. Figures~\ref{fig:appendix-d2-chronos2}--\ref{fig:appendix-d2-toto} provide the corresponding plots for the remaining three backbones, and Tables~\ref{tab:error-profile-web}--\ref{tab:error-profile-cgm} report RACE and RACE-TF outcomes, acceptance rate, and post-hoc precision for all four backbones. Test outcomes are never used to select a Gate threshold, correction, or model state.

\noindent\textbf{Figure~\ref{fig:case-study}.} The two Chronos-2 test windows are selected for qualitative illustration from a gallery stratified by the equal-weight mean of per-window MSE, MAE, and CRPS percentage improvements over the matched frozen baseline. Within each domain, windows are ranked into five equal-count groups; 20 windows are sampled from each of the first four groups and 40 from the highest-gain group. The displayed CGM case comes from the highest-gain group and the Web case from the lowest. The left column shows the query and highest-scoring RACE candidate histories before consensus filtering, followed continuously by the query future and RACE forecasts; the deployed correction aggregates the retained neighborhood, rather than using this single candidate alone. Only the last 256 steps of each 512-step history are displayed; forecasting and normalization still use the full context. The plotting widths are proportional to the displayed horizons (320:64:64), so all panels use the same physical distance per time step. Each history is standardized by its own observed-context mean and standard deviation; all target future curves use the query's statistics, with a shared vertical scale within each row. RAF retains its own retrieved input and frozen-backbone reforecast. Residual Mean uses the original unfiltered-neighborhood correction without the Gate, matching the ablation in Table~\ref{tab:ablation}. RACE-TF coincides with RACE on these two accepted windows and is therefore not plotted separately. The displayed gains use the original per-window error records, including quantile-based CRPS, while the curves show point forecasts.

\section{Resource Measurement and Material Availability}

This appendix specifies runtime measurement boundaries and the separately prepared materials for verification.

\subsection{Complete four-backbone compute ledger}

Table~\ref{tab:efficiency} reports runtime for all four backbones. These values describe measured workflows with different timing boundaries, rather than an isolated speed comparison. The following paragraphs specify the operations included for each method.

\begin{table}[t]
\caption{Recorded runtime across four frozen backbones, in wall-clock seconds (Web / CGM), with measurement boundaries. RAF includes encoding, retrieval, and augmented-input backbone inference; RACE-TF and RACE use post-cache replays, with Gate fitting and selection included for RACE. TS-RAG uses its Chronos-Bolt Base interface only; PT and FT denote pretrained and fine-tuned augmentation modules. These workflow times have different measurement boundaries and are not a matched latency benchmark.}
\label{tab:all-backbone-compute}
\label{tab:efficiency}
\begin{center}
\scriptsize
\setlength{\tabcolsep}{3.2pt}
\renewcommand{\arraystretch}{1.05}
\begin{tabular}{@{}lp{0.24\textwidth}rrrr@{}}
\toprule
Method & Measurement boundary & Chronos-2 & Chronos-Bolt Base & TimesFM-2.5 & Toto-2.0-313m \\
 &  & Web / CGM (s) & Web / CGM (s) & Web / CGM (s) & Web / CGM (s) \\
\midrule
RAF & Encoding, Top-1 retrieval, augmented-input forecasting, evaluation, and audit output. & 38.7 / 19.6 & 50.3 / 20.0 & 208.0 / 43.2 & 95.6 / 49.1 \\
RACE-TF & Post-cache route replay; residual evidence construction and one final test pass; no fitted Gate. & 91.0 / 8.7 & 93.2 / 8.4 & 92.8 / 8.5 & 93.7 / 8.7 \\
RACE & Post-cache route replay; residual evidence construction, Gate fitting, model-select, calibration, and one final test pass. & 156.3 / 19.5 & 156.7 / 19.5 & 157.2 / 19.6 & 170.8 / 20.7 \\
\midrule
TS-RAG (PT) & Frozen supplied ARM; Chronos-Bolt Base interface only. & -- & 134.2 / 9.2 & -- & -- \\
TS-RAG (FT) & Fine-tuned ARM; Chronos-Bolt Base interface only. & -- & 814.0 / 178.7 & -- & -- \\
\bottomrule
\end{tabular}

\end{center}
\end{table}

For RACE and RACE-TF, retrieval latency is measured for an already encoded query with an exact processor-based FAISS \texttt{IndexFlatL2} Top-60 index prebuilt \citep{douze2024faiss} (Web/CGM: 0.93/2.71 ms). TS-RAG uses the same exact backend on its Chronos-T5 representation with Top-10 search (1.55/3.77 ms). These query latencies exclude representation and index construction. They are separate diagnostics, not the four-backbone workflow times in Table~\ref{tab:efficiency}.

RACE and RACE-TF compute measurements use the local RTX 5070 Ti setup after frozen-backbone forecasts have been cached. They include residual evidence construction and a final test pass; RACE additionally includes Gate fitting, model selection, and calibration. They exclude backbone pretraining, forecast-cache construction, representation construction, index construction, and candidate persistence. Table~\ref{tab:efficiency} reports the independent four-backbone replay records alongside each method's measurement boundary.

RAF times come from the input-augmented runs used for Table~\ref{tab:main-results}. Timing starts after model loading and includes candidate and query encoding, exact Top-1 retrieval, augmented-input forecasting, evaluation, and audit output before the runtime record is written. These measurements include a second backbone forecast and have a different boundary from the post-cache RACE replays; the table does not support an isolated retrieval-latency or universally fastest-method claim. The TS-RAG entries retain the measured Chronos-Bolt Base pretrained-module and fine-tuned-module workflows. The supplied module's roughly one-hour pretraining on an NVIDIA A6000 \citep{ning2025tsrag} is excluded from these local measurements.

Table~\ref{tab:efficiency} combines the four-backbone runtime summary with the measurement boundary for each method. It reports Web / CGM seconds for Chronos-2, Chronos-Bolt Base, TimesFM-2.5, and Toto-2.0-313m. TS-RAG has values only for its supported Chronos-Bolt Base interface; PT and FT denote pretrained and fine-tuned augmentation modules.

\subsection{Reproducibility Package and Material Availability}

The separately prepared reproducibility package contains fixed configurations, offline environment and integrity checks, a compact result ledger, unit tests, a small public-data-derived Web fixture, and an irreversible CGM functional fixture. These fixtures exercise the complete post-forecast route---legal retrieval, residual evidence construction, Utility-Aware Correction Gate inference, and metric aggregation---while the ledger verifier recomputes the reported aggregate claims from its underlying de-identified records. As detailed in Appendix~B, the CGM fixture is a capacity-bounded stride-64, transformed-data replay; small deviations from the frozen main-paper metrics are expected and do not replace the frozen stride-32 ledgers. The fixtures are functional verification materials; they are not a replacement for the full benchmark data or an independent reproduction of every reported paper number.

The reproducibility package described here is separate from the manuscript source archive. It has been prepared for functional verification, but is not bundled with this preprint, and this version provides no public download link. It does not contain full Web data, model weights, checkpoints, raw clinical records, identifier mappings, or third-party source code. Public redistribution of the fixtures requires the applicable permissions and release terms. For a full Web run, users may invoke the documented public-asset setup utility explicitly, accepting the relevant source licenses, or supply the required data and checkpoints locally. Experiment runners do not download missing assets implicitly or silently substitute alternatives. Release of the private CGM retrieval collections remains subject to the applicable approvals and redistribution process.

\section{Limitations}

\RACE{} is designed for neighbor-rich forecasting, where a legal residual pool contains repeated or related trajectories that can provide query-specific residual evidence. Its strongest improvements therefore occur in such settings. The broader GIFT-Eval analysis shows that the procedure can remain useful when domains are more heterogeneous and neighborhood structure is less explicit, but it does not establish uniformly large gains across general forecasting collections. The results should consequently be interpreted as evidence for residual-evidence retrieval in neighbor-rich deployments, not as a guarantee of the same improvement magnitude whenever a retrieval pool is available.

The architecture is intentionally low-capacity: residual consensus is parameter-free and the learned component is a small output-side Gate. This design isolates and tests the value of legally retrieved neighbor information without adapting a TSFM backbone or relying on a backbone-specific hidden-state interface. It is not a complete study of high-capacity retrievers or correction modules, nor of large-scale retrieval libraries and broadly distributed test suites. Extending the design to larger and more diverse memory collections, evaluating richer learned residual-transfer models, and characterizing their scaling behavior are important directions for future work.

The clinical-data experiments are retrospective forecasting analyses, not validation for diagnosis, treatment recommendations, or insulin dosing.

\section{Backbone Naming Convention}
\label{app:backbone-names}

The following table defines the abbreviations used for the four evaluated backbones. M denotes million parameters; Toto2.313M refers specifically to Toto 2's 313M checkpoint. Full checkpoint identifiers remain unchanged.

\begin{center}
\small
\begin{tabular}{@{}lll@{}}
\toprule
TSFM family & Evaluated variant & Abbreviation \\
\midrule
Chronos 2 & Base & Chr.2 \\
Chronos-Bolt & Base & ChrB.B \\
TimesFM 2.5 & 200M & TFM.25 \\
Toto 2 & 313M (Toto-2.0-313m) & Toto2.313M \\
\bottomrule
\end{tabular}
\end{center}

\end{document}